\documentclass[twoside,11pt]{article}
\usepackage{jmlr2e}
\usepackage[utf8]{inputenc}
\usepackage{amsmath}
\usepackage{amsfonts}
\usepackage{mathtools}
\usepackage{url}
\usepackage{booktabs}
\usepackage{graphicx}
\usepackage{tikz}
\usepackage[vlined]{algorithm2e}
 
\usepackage{longtable}
\usepackage{enumitem}

\ShortHeadings{Surrogates of Fair Graph Filtering}{Krasanakis, and Papadopoulos}
\firstpageno{1}

\begin{document}

\title{Graph Neural Network Surrogates of Fair Graph Filtering}

\author{\name Emmanouil Krasanakis \email maniospas@iti.gr\\
       \name Symeon Papadopoulos \email papadop@iti.gr\\
       \addr 
       Information Technologies Institute\\
       Centre for Research and Technology---Hellas\\
       57001 Thermi, Thessaloniki, Greece
       }

\editor{This is an author preprint}

\maketitle

\begin{abstract}%
Graph filters that transform prior node values to posterior scores via edge propagation often support graph mining tasks affecting humans, such as recommendation and ranking. Thus, it is important to make them fair in terms of satisfying statistical parity constraints between groups of nodes (e.g., distribute score mass between genders proportionally to their representation). To achieve this while minimally perturbing the original posteriors, we introduce a filter-aware universal approximation framework for posterior objectives. This defines appropriate graph neural networks trained at runtime to be similar to filters but also locally optimize a large class of objectives, including fairness-aware ones. Experiments on a collection of 8 filters and 5 graphs show that our approach performs equally well or better than alternatives in meeting parity constraints while preserving the AUC of score-based community member recommendation and creating minimal utility loss in prior diffusion.
\end{abstract}


\section{Introduction}
Graph signal processing \citep{gavili2017shift,ortega2018graph,sandryhaila2013discrete} is a graph analysis discipline that studies node value propagation via edges. In particular, it defines graph signals as collections of prior values spread across graph nodes, such as numeric attribute values or known probabilities of nodes exhibiting a characteristic. Then, graph filters produce posterior node scores by diffusing priors through edges. Large posteriors correspond to notions of structural proximity to nodes with large priors. 

This scheme facilitates many downstream graph mining tasks, such as unsupervised extraction of tightly knit node clusters \citep{andersen2006local,schaeffer2007graph,kulis2009semi,wu2012learning}, node recommendation based on structurally proximity to a set of query nodes \citep{tong2006fast,kloster2014heat}, and certain graph neural network architectures for node or graph classification \citep{kipf2016semi,chen2020simple,klicpera2018predict,dong2020equivalence,huang2020combining}. 
As a running example, starting from a set of a social network's users that have formed a community (social group) based on a shared interest, filters can recommend the community to other users based on the structure of the social network's interaction graph.

\par
Fairness concerns arise when the outputs of artificial intelligence and data mining systems are correlated to protected attribute values, such as gender or ethnicity \citep{chouldechova2017fair,kleinberg2018algorithmic}.\footnote{We focus on binary protected attributes in that they obtain values either 1 or 0 for each data sample.} In these cases, one bias mitigation goal is to consider organization of data samples into protected and non-protected groups depending on given attribute values (e.g., men and women) and require that both groups should produce similar outputs when evaluated by the same measures of choice \citep{chouldechova2017fair,krasanakis2018adaptive,zafar2019fairness,ntoutsi2020bias}. A popular fairness objective is the statistical parity of the fractions of positive prediction between the groups; this is known as disparate impact elimination \citep{biddle2006adverse,calders2010three,kamiran2012data,feldman2015certifying} and often assessed through a measure called \emph{prule} (Subsection~\ref{fairness background}). 
\par
Fairness concerns also arise for graph filters, whose posteriors can be affected by prior value or graph structure biases (Subsection~\ref{fairness background}). For example, the average posterior score for men could be higher than the average score for women if nodes with large priors were mostly men and the graph featured few inter-gender interactions. To quantify how fair graph filters are with respect to disparate impact, we employ a generalization of the prule that accounts for node scores \citep{krasanakis2020applying} and aim to maximize it 
while in large part maintaining the original posteriors. This way, we can retain the predictive capabilities of graph filters within downstream tasks, such as the AUC of recommendations (Subsection~\ref{community}), while addressing disparate impact concerns.
\par
To achieve this, we expand on a previous hypothesis \citep{krasanakis2020applying} that graph filter fairness can be induced by editing priors so that resulting posteriors largely preserve the outcome of filtering but also optimize fairness-aware objectives. Here, we introduce a novel mathematical framework to theoretically ground this hypothesis, as well as a drastically improved mechanism for editing posteriors that accounts for our new analysis (Section~\ref{optimizing posterior scores}). We also reframe this approach as a type of graph neural network that parses as inputs the original prior values, the original posteriors, and the sensitive attribute, and is used as a surrogate model of ideal posteriors (Section~\ref{surrogate posteriors}). The network's neural parameters are learned on-the-fly depending on the original priors and filter, and architectural hyperparameters are chosen so that they satisfy conditions stipulated by our analysis.
\par
The contribution of this work is threefold:
\begin{itemize}[noitemsep,topsep=0pt]
\item[a)] We introduce a novel graph neural network framework for adjusting the outcome of graph filtering so that it becomes fairness-aware. This substitutes most filters within existing systems by maintaining the same notions of structural proximity while making sure that groups of nodes are protected.
\item[b)] We prove that prior editing can control posteriors to reach local optima in many objectives, such as fairness-aware ones. We also show that appropriate L1 regularization can make any twice differentiable objective suitable to prior editing. Universal approximation via filtering is also applicable to objectives other than ours.
\item[c)] We compare our approach to existing ones on a corpus of 5 multidisciplinary real-world graphs combined with 8 graph filters and variations of 2 downstream tasks (community member recommendation and node value diffusion) to show that the proposed framework better preserves posteriors while mitigating their disparate impact. 
\end{itemize}
\par

\section{Background and Related Work}\label{background}
In this section we provide the theoretical background necessary to understand our work. First, in Subsection~\ref{gsp} we overview graph signal processing concepts used to study a wide range of methods for obtaining node scores given prior information of node values. In Subsection~\ref{community} we also present two popular graph filtering tasks, namely community member recommendation, and node score diffusion. These are employed by many practical applications, and we later use them as the main scenarios for evaluating fairness-aware graph filtering. Additionally, in Subsection~\ref{fairness background} we discuss algorithmic fairness under the prism of graph signal processing, and overview the limited research done to merge these disciplines. Finally, Subsection~\ref{gnns} introduces related graph neural network principles and terminology. The operations and symbols defined in this work are summarized in Table~\ref{notation}.

\begin{table}[hbt]
\centering
\footnotesize
\label{notation}
\begin{tabular}{l|l}
    \textbf{Notation} & \textbf{Interpretation}\\
    \hline 
        $\mathcal{I}$ & Identity matrix with appropriate dimensions\\
        $\textbf{0}$ & Column vector of appropriate rows and zero elements\\
        $r[v]$ & Element corresponding to node $v$ of graph signal $r$\\
        $\mathcal{L}(r)$ & Loss function for graph filter posteriors $r$\\
        $\phi$ & The parameter controlling $\phi$-fairness definitions\\
        $\theta$ & Parameters (typically neural ones) of prior editing/generation schemes\\
        $\mathcal{M}(\theta)$ & Prior graph signal generation scheme\\
        $\nabla \mathcal{L}(r)$ & Gradient vector of loss $\mathcal{L}(r)$ with elements $\nabla \mathcal{L}(r)[v]=\frac{\partial \mathcal{L}(r)}{\partial r[v]}$\\
        $\mathbb{J}_{\mathcal{M}}(\theta)$ & The Jacobian matrix of multivariate function $\mathcal{M}(\theta)$ with $\mathbb{J}_\mathcal{M}(\theta)[v]=\nabla_\theta (\mathcal{M}(\theta)[v])$ for rows $v$\\
        $\mathbb{H}_{\mathcal{L}}(r)$ & The Hessian matrix of a real-valued function $\mathcal{L}(r)$ obtained per $\mathbb{H}_{\mathcal{L}}(r)=\mathcal{\nabla \mathcal{L}}(r)$\\
        $|x|$ & Absolute value for numbers, number of elements for sets\\
        $\|x\|$ & L2 norm of vector $x$ computed as $\sqrt{\sum_v x[v]^2}$\\
        $\|x\|_1$ & L1 norm of vector $x$ computed as $\sum_v |x[v]|$\\
        $\|x\|_\infty$ & Maximum value of $x$ computed as $\max_v x[v]$\\
        $\lambda_1$ & Smallest eigenvalue of a positive definite matrix\\
        $\lambda_{\max}$ & Largest eigenvalue of a positive definite matrix\\
        $\hat{A}$ & Normalized version of adjacency matrix $A$\\
        $F(\hat{A})$ & Graph filter on normalized adjacency matrix $\hat{A}$\\
        $A\setminus B$ & Set difference, that is the elements of $A$ not found in $B$\\
        $a^T b$ & Dot product of column vectors $a,b$ as matrix multiplication\\
        $\mathbb{R}^{|\mathcal{V}|}$ & Space of column vectors comprising all graph nodes\\
        $\text{diag}([\lambda_i]_i)$ & A diagonal matrix $\text{diag}([\lambda_i]_i)[i,j]=\{\lambda_i\text{ if }i=j,0\text{ otherwise}\}$\\
        $A[u,v]$ & Element of matrix $A$ at row $u$ and column $v$\\
        $A^T$ & Transposition of matrix $A$ for which $A^T[u.v]=A[v,u]$\\
        $A^{-1}$ & Inverse of invertible matrix $A$\\
        $\{x\,|\,cond(x)\}$ & Elements $x$ satisfying a condition $cond$\\
        $\mathcal{S}$ & Set of nodes with sensitive attribute values\\
\end{tabular}
\caption{Mathematical notation. Graph-related quantities refer to a common studied graph.}
\end{table}

\subsection{Graph Signal Processing}\label{gsp}
\subsubsection{Graph signal propagation} 
Graph signal processing \citep{gavili2017shift,ortega2018graph,sandryhaila2013discrete} is a domain that extends traditional signal processing to graph-structured data. To do this, it starts by defining graph signals $q:\mathcal{V}\to\mathbb{R}$ as maps that assign real values $q[v]$ to graph nodes $v\in\mathcal{V}$.\footnote{\footnotesize Signals with multidimensional node values can be expressed as ordered collections of real-valued signals.} Graph signals can be represented as column vectors $q'\in\mathbb{R}^{|\mathcal{V}|}$ with elements $q'[i]=q[\mathcal{V}[i]]$, where $|\cdot|$ is the number of set elements and $\mathcal{V}[i]$ is the $i$-th node of the graph after assuming an arbitrary fixed order. For ease of notation, in this work we use graph signals and their vector representations interchangeably by replacing nodes with their ordinality index. In other words, we assume the isomorphism $\mathcal{V}[i]= i$. Intuitive interpretations of information captured by graph signals are presented in Subsection~\ref{community}.
\par
A pivotal operation in graph signal processing is the one-hop propagation of node values to their graph neighbors, where incoming values are aggregated on each node. Expressing this operation for unweighted graphs with edges $\mathcal{E}\subseteq\mathcal{V}\times\mathcal{V}$ requires the definition of adjacency matrices $A$, whose elements correspond to the binary existence of respective edges, i.e., $A[i,j]=\{1\text{ if }(\mathcal{V}[i], \mathcal{V}[j])\in\mathcal{E}, 0\text{ otherwise}\}$. A normalization operation is typically employed to transform adjacency matrices into new ones $\hat{A}$ with the same dimensions, but which model some additional assumptions about the propagation mechanism (see below for details). Then, single-hop propagation of node values stored in graph signals $q$ to neighbors yields new graph signals $q_{next}$ with elements $q_{next}[u]=\sum_{v\in\mathcal{V}}\hat{A}[u,v]q[v]$. For the sake of brevity, this operation is usually expressed using linear algebra as $q_{next}=\hat{A}q$.
\par
Two popular types of adjacency matrix normalization are column-wise and symmetric. The first views one-hop propagation as a stochastic process \citep{tong2006fast} that is equivalent to randomly walking the graph and selecting the next node to move to from a uniformly random selection between neighbors. Formally, this is expressed as $\hat{A}_{col}=AD^{-1}$, where $D=\text{diag}\big(\big[\sum_{v}A[u,v]\big]_{u}\big)$ is the diagonal matrix of node degrees. Columns of the column-normalized adjacency matrix sum to $1$. This way, if graph signal priors model probability distributions over nodes, i.e., their values sum to $1$ and are non-negative, posteriors also model probability distributions.
\par
On the other hand, symmetric normalization arises from a perspective where the eigenvalues of the normalized adjacency matrix are treated as the graph's spectrum \citep{chung1997spectral,spielman2012spectral}. In this case---and if the graph is unweighted and undirected in that the existence of edges $(u,v)$ also implies the existence of edges $(v,u)$---a symmetric normalization formula is needed to guarantee that eigenvalues are real numbers. The normalization $\hat{A}_{symm}=D^{-1/2}AD^{-1/2}$ is predominantly selected, on merit that it has bounded eigenvalues (see below) and $D^{1/2}(I-\hat{A}_{symm})D^{1/2}$ is a Laplacian operator that implements the equivalent of discrete derivation over graph edges.
\par
Graph signal processing research often targets undirected graphs and symmetric normalization, since these enable computational tractability and closed form convergence bounds of resulting tools.\footnote{\footnotesize Theoretical groundwork has also been established to consider spectral equivalent for undirected graphs and, ultimately, asymmetric adjacency matrix normalization \citep{chung2005laplacians,yoshida2019cheeger}. However, extending our analysis to these is non-trivial.} In this work, we also favor this practice, because it allows graph signal filters to maintain spectral characteristics needed by our analysis. The key property we take advantage of is that, as long as the graph is connected, the normalized adjacency matrix $\hat{A}$ is invertible and has the same number of real-valued eigenvalues as the number of graph nodes $|\mathcal{V}|$. Furthermore, all eigenvalues reside in the range $[-1,1]$. If we annotate these eigenvalues as  $\{\lambda_i\in[-1,1]\,|\,i=1,\dots,|\mathcal{V}|\}$, the symmetric normalized adjacency matrix's Jordan decomposition takes the form: 
\begin{equation*}
    \hat{A}=U^{-1} \text{diag}([\lambda_i]_i)U
\end{equation*} 
where $U$ is an orthonormal matrix with columns the corresponding eigenvectors. Therefore, scalar multiplication, power and addition operations on the adjacency matrices also transform eigenvalues the same way. For example, it holds that $\hat{A}^n=U^{-1}\text{diag}([\lambda_i^n]_i)U$.
\subsubsection{Graph filters} The one-hop propagation of graph signals is a type of shift operator in the multidimentional space induced by the graph, in that it propagates values based on a notion of structural/relational proximity. Based on this observation, graph signal processing uses it analogously to the time shift $z^{-1}$ operator and defines the notion of graph filters as weighted aggregations of multi-hop propagation. In particular, since $\hat{A}^nq$ expresses the propagation $n=0,1,2,\dots$ hops away of graph signals $q$, a weighted aggregation of these hops means that the outcome of graph filters can be expressed as:
\begin{equation}\label{filter}
\begin{split}
&r=F(\hat{A})q\\
&F(\hat{A})=\sum_{n=0}^\infty f_n \hat{A}^n
\end{split}
\end{equation}
where $F(\hat{A})$ is the filter characterized by real-valued weights $\{f_n\in\mathbb{R}|n=0,1,2,\dots\}$ indicating the importance placed on propagation of graph signals $n$ hops away. In this work, we understand the resulting graph signal $r$ to capture {posterior node scores} that arise from passing {prior node values} of the original signal $q$ through the filter. Given the weighted aggregation of multi-hop propagations, a generalized understanding of posteriors is in terms of quantifying proximity to nodes with large prior values via many paths of length $n$ for lengths corresponding to large weights $f_n$. Different interpretations of which real-world properties are captured by proximity give rise to different graph mining tasks, such as the ones presented in Subsection~\ref{community}.
\par
In this work, we focus only on graph filters that are positive definite matrices, i.e., whose eigenvalues are all positive. For symmetric normalized graph adjacency matrices with eigenvalues $\{\lambda_i\in[-1,1]|i=1,\dots,|\mathcal{V}|\}$, it is easy to check whether graph filters defined per Equation~\ref{filter} are positive definite, as they assume eigenvalues $\big\{F(\lambda_i)=\sum_{n=0}^\infty f_n\lambda_i^n\,|\,i=1,\dots,|\mathcal{V}|\big\}$
and hence we can check whether the graph filter's corresponding polynomial assumes only positive values:
\begin{equation}\label{condition}
    F(\lambda)>0\,\forall \lambda\in[-1,1]
\end{equation}
For example, filters arising from decreasing importance of propagating more hops away $f_n>f_{n+1}>0\,\forall n=0,1,\dots$ are positive definite.  
Two well-known graph filters that are positive definite for symmetric adjacency matrix normalizations are personalized pagerank \citep{andersen2007local,bahmani2010fast} and heat kernels \citep{kloster2014heat}. These respectively arise from power degradation of hop weights $f_n=(1-a)a^n$ and the exponential kernel $f_n=e^{-t}{t^n}/{n!}$ for parameters $a\in[0,1]$ and $t\in\{1,2,3,\dots\}$. The best parameter values vary, depending on the graph and mining task.

\subsection{Example Applications of Graph Filtering}\label{community}
\subsubsection{Community member recommendations}
Nodes of real-world graphs are often organized into communities of either ground truth structural characteristics \citep{fortunato2016community,leskovec2010empirical,xie2013overlapping,papadopoulos2012community} or shared node attribute values \citep{hric2014community,hric2016network,peel2017ground}. A common task in graph analysis is to score or subsequently rank all nodes based on their relevance to such communities. This is particularly important for large graphs, where community boundaries can be vague \citep{leskovec2009community,lancichinetti2009detecting}. Furthermore, node scores can be combined with other characteristics, such as their past values when discovering nodes of emerging importance in time-evolving graphs, in which case they should be of high quality across the whole graph. Many algorithms that discover communities from only a few known members also rely on transforming and thresholding node scores \citep{andersen2006local,whang2016overlapping}.
\par
In many applications, node scores are computed with graph filters. In particular, prior graph signals are constructed given the knowledge that certain example nodes exhibit a common attribute of interest. This knowledge may be inputted from real-world data gathering or, in case of unsupervised community detection, by automatic extraction of candidate nodes around which communities should be discovered. In both cases, prior signal elements are assigned binary values depending on whether respective nodes are provided as examples: 
\begin{equation*}
q[v]=\{1\text{ if }v\text{ is known to have the attribute}, 0\text{ otherwise}\}
\end{equation*}

In general, some but not all nodes that belong to the same structural community could serve as examples to construct a prior graph signal. All known members should be used, but there may be unknown ones too. 
Then, graph filters $F(\hat{A})$ can be deployed to produce posteriors $r=F(\hat{A})q$ whose elements $r[v]$ indicate the structural proximity of nodes $v$ to known members. Finally, nodes with large posteriors are good candidates to be recommended as community members, to be granularly identified as exhibiting the shared metadata attribute of the examples, or, depending on the application, to receive recommendations about the studied community. The success of such actions depends on the choice of filter \citep{krasanakis2022autogf}, which can be determined either with an informed identification of dominant structural characteristics leading to the creation of the community, or with supervised experimentation on domain graphs.
\par
The measure that typically quantifies the quality of community member recommendations arising from node scores is the area under the curve of receiver operating characteristics (AUC) \citep{hanley1982meaning}, which compares operating characteristic trade-offs at different decision thresholds. If we pack ground truth node scores in a graph signal $q_{test}[v]=\{1\text{ if node }v\text{ is a community member},0\text{ otherwise}\}$, so as to evaluate a posterior score signal $r[v]$, the True Positive Rate (TPR) and False Positive Rate (FPR) operating characteristics for decision thresholds $\theta$ can be respectively defined as:
\begin{align*}
   &TPR(\theta)=P(q_{test}[v]=1\,|\,r[v]\geq\theta)\\
   &FPR(\theta)=P(q_{test}[v]=0\,|\,r[v]\geq\theta)
\end{align*}
where $P(a|b)$ denotes the probability of event $a$ conditioned on $b$. AUC is defined as the cumulative effect induced to the TPR when new decision thresholds are used to change the FPR and quantified per the following formula:
\begin{equation}\label{AUC}
    AUC=\int_{-\infty}^\infty TPR(\theta) FPR'(\theta)\,d\theta
\end{equation}
AUC values closer to $100\%$ indicate that community members exhibit higher scores compared to non-community members, whereas $50\%$ AUC corresponds to random node scores.

\subsubsection{Graph diffusion}
Graph diffusion refers to procedures that smooth prior node values throughout graph structures. This task directly translates to a graph signal processing pipeline, where node values are organized into prior graph signals and graph filters play the role of the diffusion mechanism. The notion of structural proximity is understood as an indication of prior-posterior correlations between nodes. Based on this understanding, diffusion has been used for inference of continuous-valued node attributes when nodes exhibit missing or noisy information \citep{castillo2007know,gadde2013bilateral,anis2016efficient}. These efforts have culminated in integrating diffusion mechanisms in graph neural networks that smooth the prediction logits of multilayer perceptrons through the graph structure (Subsection~\ref{gnns}).

Compared to community member recommendation, which typically handles binary priors indicating community membership---and therefore may start from only a few non-zero prior values---diffusion parses continuous (sometimes even negative) values that could be non-zero for most graph nodes. One could be tempted to think of diffusion as the more general setting, but in practice the two domains often remain distinct due to addressing different application needs that present different engineering and theoretical challenges. For example, when looked under the prism of graph filters, diffusion can perfectly reconstruct \citep{anis2016efficient} appropriately downsampled graph signals \citep{anis2016efficient}, but community analysis frequently dabbles into extremely sparse data that produce large posteriors in (and can hence be restricted to run on) only certain graph segments \citep{lofgren2016personalized}.

There are many branching applications of diffusion where its efficacy is tangled with other application domain intricacies, such as system hyperparameters. Thus, it has been implicitly acknowledged \citep{tsioutsiouliklis2020fairness,tsioutsiouliklis2021fairness} that exploratory variations involving fairness-aware objectives should not be evaluated in a case-by-case basis but in a vacuum, where regression error measures capture the distance between the original (biased) posteriors and debiased ones. In this work, we follow the same principle and leave for future work the application of fair diffusion to specific downstream tasks.

We thus quantify the impact of inserting fairness postulates into diffusion mechanisms by measuring the error between graph filter posteriors $r_{original}$ and the fairness-aware ones $r_{fair}$ generated by ours or other approaches. In the literature, the absolute error commonly serves this purpose, but we point out that it largely ignores the impact on lower-scored posteriors, such as sensitive group members. To prevent the utility loss itself from suffering from disparate impact, and given that we later experiment in settings that yield positive posteriors, in this work we work with the average relative error as the utility loss:
\begin{equation}\label{utility}\mathcal{L}_{util}=\tfrac{1}{|\mathcal{V}|}{\big\|1-\tfrac{r_{fair}}{r_{original}}\big\|_1}\end{equation}
Utility losses should ideally be small, but for biased original posteriors there inevitably exist trade-offs between zero losses and deviations from trying to make new posteriors fair.

\subsection{Graph Neural Networks}\label{gnns}
Graph neural networks generalize base graph diffusion principles, which are applied to graph signals with numeric node values, to multidimensional node feature values. Given node feature matrices $H^{(0)}$ whose columns can be considered graph signals of respective features (their rows correspond to node feature values) graph neural networks are broadly derived by adjusting the following equation \citep{kipf2016semi}:
\begin{equation*}
    H^{(\ell)}=\sigma\big(\hat{A}H^{(\ell-1)}W^{(\ell)}+b^{(\ell)}\big)
\end{equation*}
where $\hat{A}$ is the symmetrically normalized adjacency matrix (often edited with the renormalization trick of adding self-loops in the graph) and $W^{(\ell)}$ and $b^{(\ell)}$ define a dense affine layer transformation of the graph shift $\hat{A}H^{(\ell-1)}$. The output of each transformed graph shift passes through a non-linear transformation $\sigma$, which at intermediate layers is typically chosen as the rectified linear activation $\text{relu}(x)=max\{0,x\}$ applied element-wise. When graph neural networks are used for node classification, their output is activated through a row-wise softmax to approximate one-hot encoding of predicted classes. In this work, our output activation and objective are rather different.
\par
Recent research has shown that the above type of architecture suffers from oversmoothing and should thus be limited to a couple of layers that fail to account for node information many hops away. To address this issue, recursive schemes have been proposed to partially maintain early representations on later layers. Of these, our work resembles the predict-then-propagate architecture \citep{klicpera2018predict}, which decouples the neural and graph diffusion components by first enlisting multilayer perceptrons to end-to-end train representations $H^{(L)}$ of node features, and then applying graph filters on each representation dimension to arrive at final predictions:
\begin{align*}
&H^{(\ell)}=\text{relu}(H^{(\ell-1)}W^{(\ell)}+b^{(\ell)})\quad\ell=1,2,\dots,L\\
&\hat{Y}_{logits}=F(\hat{A})H^{(L)}\\
&\hat{Y}=\text{softmax}(\hat{Y}_{logits})
\end{align*}
Originally, this approach used personalized pagerank as the filter of choice $F(\hat{A})$.
\par
Our analysis arrives at a similar scheme of producing predictive logits $\hat{Y}_{logits}$, but finds that the selection of graph filter should depend on the desired objective in that it should yield qualitative priors to be adjusted. The applied setting also differs from the typical graph neural network formulation in that we extract node features $H^{(0)}$ from the graph filtering process and directly use the logits as surrogate estimators of ideal posteriors optimizing fairness-aware objectives. Furthermore, our approach is designed only for objectives that evaluate one-dimensional node ranking. Prospective extension to scopes like node classification is outlined in Appendix~\ref{towards universal} but requires non-trivial theoretical groundwork to extend our analysis, as well as experimental validation that is left to future work.

\subsection{Fairness of graph filter posteriors}\label{fairness background}
In this subsection we introduce the well-known concept of disparate impact in the domain of algorithmic fairness, which we aim to mitigate. We also explore systemic graph biases and overview previous works that inject fairness in graph filter posteriors.

\subsubsection{Disparate impact elimination} When not focused on equal treatment between specific individuals, algorithmic fairness is broadly understood as parity between sensitive and non-sensitive samples over a chosen statistical property. Three popular fairness-aware objectives \citep{chouldechova2017fair,krasanakis2018adaptive,zafar2019fairness,ntoutsi2020bias} are disparate treatment elimination, disparate impact elimination, and disparate mistreatment elimination. These correspond to not using the protected attribute in predictions, preserving statistical parity between the fraction of sensitive and non-sensitive positive labels for sample groups with the same protected attribute values, and achieving identical predictive performance on the two groups under a measure of choice.
\par
Here, we focus on mitigating disparate impact unfairness \citep{chouldechova2017fair,biddle2006adverse,calders2010three,kamiran2012data,feldman2015certifying}. An established measure that quantifies this objective for binary protected attributes is the \textit{prule} \citep{biddle2006adverse}. Denoting as $R[v]\in\{0,1\}$ the binary outputs of a system $R$ for samples $v$, $S$ the group (set) of protected samples and $S'$ its complement, this measure is defined as:
\begin{equation}
\begin{split}
    &prule = \frac{\min\{r_{S},r_{S'}\}}{\max\{r_{S},r_{S'}\}}\in[0,1]\\
    &r_{S} = P(R[v]=1|p\in S)\\
    &r_{S'} = P(R[v]=1|p\not\in S)
\end{split}
\end{equation}
The prule computes to 100\% when the fractions $r_S,r_{S'}$ are perfectly equal and to 0\% when there is no positive output for one of the sensitive attribute's values. There is precedence \citep{biddle2006adverse} for considering $80\%$ prule or higher fair.
\par
Calders-Verwer disparity $|r_S-r_{S'}|$ \citep{calders2010three} is also a well-known disparate impact assessment measure. However, although it is optimized at the same point as the prule, it biases fairness assessment against high fractions of positive predictions. For example, it considers the fractions of positive labels $(r_S,r_{S'})=(0.8, 0.6)$ less fair than $(r_S,r_{S'})=(0.4, 0.3)$. We shy away from this understanding because we later employ a stochastic interpretation of posterior nodes scores that could be scaled by a constant yet unknown factor. On the other hand, the prule would quantify both numerical examples as $\tfrac{0.6}{0.8}=\tfrac{0.3}{0.4}=75\%$ fair. 

\subsubsection{Posterior biases in graphs}\label{posterior biases}
Fairness concerns also arise in graph filters. To begin with, linked social network nodes often exhibit similar attribute values. This phenomenon is known as homophily \citep{mcpherson2001birds} and is the core argument behind the use of low-pass graph filters, i.e., which concentrate on propagating priors only a few hops away. However, the same phenomenon also means that prior biases (e.g., lower values on average against protected group nodes) are transferred to the posteriors of low-pass graph filters, such as personalized pagerank \citep{karimi2018homophily}. 
Additionally, there may exist structural biases in graphs with non-homogenous node degree distributions \citep{vaccario2017quantifying,cui2021algorithmic}. For example, graph filters often assign higher scores to well-linked nodes (as \cite{fortunato2006approximating} demonstrate for the personalized pagerank filter) and this causes those sensitive attribute values that are correlated with node degrees to also be correlated with node scores.
\par
\begin{figure}[htbp]
\centering
    \includegraphics[width=0.35\textwidth,clip]{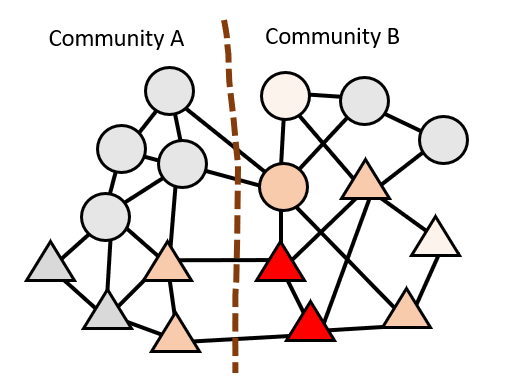}
    \includegraphics[width=0.35\textwidth,clip]{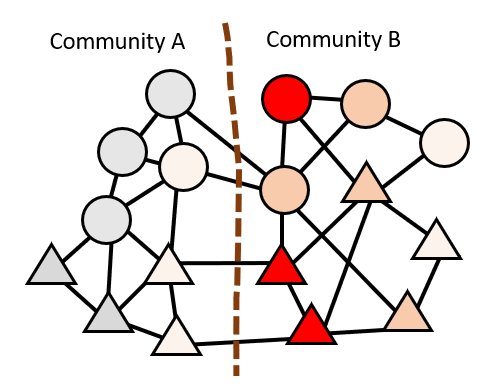}
    \caption{Biased (left) and fair (right) community member recommendations with regards to the protected subgroup of circles.}
    \label{fig:biased_filter}
\end{figure}

As an example of how graph filtering can be biased, consider the left graph of Figure~\ref{fig:biased_filter} and the node shape as a sensitive attribute. The circle and triangle groups of nodes tend to form more intra-group than other edges, but there also exist two structural communities A and B. Constructing graph signal priors from the red known members of community B, which are both triangles, leads an example graph filter towards favoring member recommendations (pink) for more triangles rather than the top circles, even if the former happen to reside in community A instead. These recommendations are in line with notions of structural proximity, but if the goal was to recommend members of community B, example members and respective priors are clearly biased in that they do not represent any circles and thus fail to indicate the top-right area of the graph as important. One way to mitigate this type of bias would be by including circles in the examples, as shown on the right graph of the same figure. 

Prior bias could come from unfair data gathering procedures. However, given a small number of example community members, this bias can also arise inadvertently, due to random selection of structurally close same-group nodes as examples. This is unlikely for unbiased sampling mechanisms and large numbers of example nodes, but it is commonplace to perform graph filtering with as few as one or two non-zero prior values. Regardless of its cause (e.g., chance or malice), prior bias can be exacerbated by homophilous graph structures. One view of this phenomenon is that the structures themselves are biased \citep{karimi2018homophily}. However, in this work we show that appropriate graph signal prior selection can mitigate any structural biases in connected graphs, which indicates that bias can also be seen as the fault of priors, namely of large prior values residing structurally far away from nodes of the protected group.

\subsubsection{Posterior score fairness}
In domains related to the outcome of graph mining algorithms, fairness has been defined for the order of recommended items \citep{beutel2019fairness,biega2018equity,yang2017measuring,zehlike2017fa} as equity in the ranking positions between protected and non-protected group members. These notions of fairness are not applicable to the more granular understanding provided by node scores.
\par
Another definition of graph mining fairness has been introduced for node embeddings \citep{bose2019compositional,rahman2019fairwalk} under the guise of fair random walks, which is the stochastic process modeled by personalized pagerank when the adjacency matrix is normalized by columns. Yet, the fairness of these walks is only implicitly asserted through embedding fairness. Furthermore, they require at least one member of the protected group to be adjacent to each node, to make sure that the walks can favor them at every step on which fairness needs to be corrected. 
\par
Fairness has also been recently explored for the outcomes of graph neural networks \citep{dai2020fairgnn,ma2021subgroup,dong2021individual,dai2021say,zhang2022fairness,dong2022fairness}. Approaches are typically trained to produce fair predictions (e.g., recommendations, link predictions) by a) incorporating traditional notions of fairness as regularizers to differentiable relaxations of accurate node classification (e.g., cross-entropy minimization), b) imposing similar fairness constraints on the training process, or c) rebalancing algorithmic components to debias the inference process. These works also support partial knowledge of sensitive attribute values, given that these can be neurally estimated. In line with fairness-aware graph filtering (see below), a popular type of rebalancing constitutes of adding or removing edges \citep{loveland2022fairedit}.

Despite the existence of works tailored to other graph mining paradigms, it remains important to investigate fairness for traditional graph filtering. To begin with, advances on graph neural network theory \citep{dong2020equivalence} suggest that this paradigm can be equivalent to decoupling neural network and graph filter components. Thus, the question of how to make the outcome of graph filters fair can also be of use there. At the same time, graph neural networks do not see use in all graph mining applications, as they tend to rely on an abundance of node features, which are not always present for graph mining, and are often computationally intractable to parse with modern GPUs at the scale of graphs with millions of nodes. Finally, existing architectures tend to focus on node classification or link prediction instead of the more granular scoring we explore in this work.
\par
Recent works by \cite{tsioutsiouliklis2020fairness,tsioutsiouliklis2021fairness,tsioutsiouliklis2022link} have initiated a discourse on the posterior score fairness of graph filters by recognizing the need for optimizing trade-offs between fairness and posterior preservation. Furthermore, they provide a first definition of node score fairness, called $\phi$-fairness, which for scores $r[v]$ of graph nodes $v\in\mathcal{V}$ allocates a fixed fraction of their total mass for the protected group nodes $v\in S$:
\begin{equation*}
   \sum_{v\in S} r[v]=\phi \sum_{v\in \mathcal{V}} r[v]
\end{equation*}
To achieve perfect compliance to $\phi$-fairness, the same researchers apply graph structure modifications (edge addition or removal) to balance the score mass flowing to and from protected group nodes during graph shifts. Furthermore, they address fairness-aware algorithms for personalized graph filter inputs by introducing the concept of \textit{universal fairness} as the practice of modifying the graph structure so that $\phi$-fairness is satisfied irrespective of provided priors. They also show that graph modifications are the only way to make the personalized pagerank filter universally fair.

Unfortunately, modifying graphs so that filters become universally fair admits several restrictions that are not always possible to satisfy. First, this kind of approach has been analysed only for the family of personalized pagerank filters with column-wise normalization; modifications could exist for symmetrically normalized personalized pagerank or other filters, but additional case-by-case analysis is needed. Second, not only are the modifications that would best preserve the original posteriors still unknown, but existing mechanisms remain dependent on priors. As an extension of this point, we argue that the intent to make algorithms universally fair---even for graph signal priors that capture biases---introduces drastic modifications in what graph filters consider as structural proximity and ultimately hinders the preservation of original posteriors by affecting them more than needed. In fact, some modifications artificially penalize the scores of nodes with non-zero priors to offset some of the biased probability mass retained by personalized pagerank.

\par
A different proposition for fairness-aware graph filtering is to tailor bias mitigation to the specific priors injecting unfairness \citep{krasanakis2020applying}. In this work, we expand on this direction and aim to improve upon preliminary heuristics of editing priors that optimize fair graph filter posterior objectives without altering the relations between nodes. This approach addresses traditional disparate impact mitigation by generalizing the prule as a measure of posterior score fairness. The generalization starts from a stochastic interpretation of posterior node scores, where they are proportional to the probability of nodes assuming positive labels, and calculates the expected positive number of sensitive and non-sensitive node labels obtained by sampling mechanism that uniformly assigns positive labels with probabilities proportional to posterior scores:
\begin{equation*}
    \begin{split}
        &p_S=P(R[v]=1|p\in S)=\tfrac{\Omega}{|S|} {\sum_{v\in S} r[v]}\\
        &p_{S'}=P(R[v]=1|p\not\in S)=\tfrac{\Omega}{|\mathcal{V}\setminus S|}{\sum_{v\not\in S} r[v]}
    \end{split}
\end{equation*}
where the value $\Omega>0$ is the scale of the proportion and $R$ is a stochastic process with probabilities $P(R[v]=1)=\Omega\, r[v]$. Plugging these in the prule cancels out the scale between the nominator and denominator and yields the following formula for calculating a stochastic interpretation of the prule given posterior node scores $r$:
\begin{equation}\label{prule}
\begin{split}
    &prule = \frac{\min\big\{|\mathcal{V}\setminus S|\sum_{v\in S}r[v],|S|\sum_{v\not\in S}r[v]\big\}}{\max\big\{|\mathcal{V}\setminus S|\sum_{v\in S}r[v],|S|\sum_{v\not\in S}r[v]\big\}}
\end{split}
\end{equation}
We also adopt this approach in this work and, by convention, consider $prule=0$ when all node scores are zero. Although conceived through independent processes, the conditions of 100\% prule and $\phi$-fairness capture the same type of posterior equity when $\phi=\tfrac{|S|}{|\mathcal{V}|}$, i.e., they become equivalent in this case. When the protected group nodes end up with smaller scores on average, which is the typical case we address in this work, the relation between the prule and $\phi$ generalizes to:
$$prule=\frac{\phi|\mathcal{V}\setminus S|}{(1-\phi)|S|}\Leftrightarrow \phi=\frac{|S|prule}{|\mathcal{V}|+|S|(prule-1)}$$
Otherwise, the desired correspondence between the two notions of fairness can be found by considering the complement $\mathcal{V}\setminus S$ as the protected group.

\section{Prior Editing to Optimize Posterior Scores}\label{optimizing posterior scores}
In this work we attempt to make graph filtering fairness-aware while respecting how it understands structural proximity. Before working on the fairness domain, in this section we conduct a more general analysis of how to search for posteriors optimizing a broad class of objectives while respecting the propagation mechanisms of graph filters. Our proposed theoretical framework is exploited later on, in Section~\ref{surrogate posteriors}, to eliminate disparate impact from node scores while minimally perturbing the outcome of traditional graph filtering.

One possible approach to optimizing posteriors would be by directly adjusting them \citep{tsioutsiouliklis2020fairness}. However, this practice deteriorates the quality gained by passing priors through graph filters. Ultimately, this loses the robustness of propagating information through node relations by introducing a degree of freedom for each node (Subsection~\ref{threats to applicability}). For example, under this approach, there is no difference in increasing the scores of low-scored nodes in the protected group by a small but non-negligible amount compared to increasing the scores of high-scored nodes of the same group. However, the first case could lead to a catastrophic loss of intra-group recommendation order, even if it were pareto optimal with respect to any small node score permutation. 

\begin{figure}[htbp]
\centering
    \includegraphics[width=0.75\textwidth,clip]{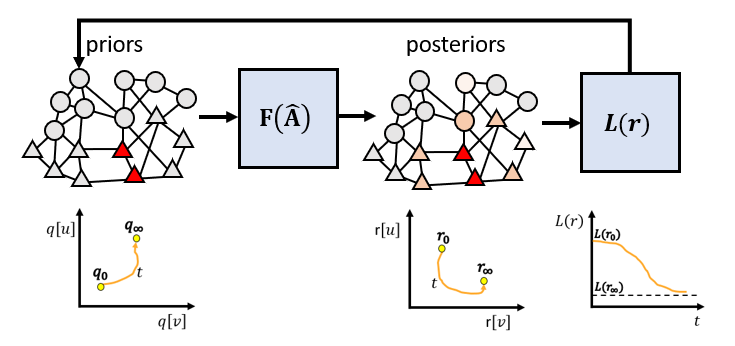}
    \caption{A prior editing process and its optimization trajectory for two nodes.}
    \label{fig:overview1}
\end{figure}

To prevent loss of posterior quality gained by passing graph signal priors through filters, for instance to facilitate downstream tasks, it has been proposed that, instead of the posteriors, the priors should be edited \citep{krasanakis2020applying}. In this work, we theoretically back this claim by deriving the editing as the convergent point of a gradual procedure that leads to posterior objective optimization near the original priors. An intuitive understanding of this process is presented in Figure~\ref{fig:overview1}, where original priors $q_0$ are edited over time $t$ until they reach new ones $q_\infty$. The corresponding original posteriors $r_0$ turn into new node scores $r_\infty$ that asymptotically reach the local optima of an objective $\mathcal{L}(r_\infty)$.

Individually editing each prior node value may still introduce too many degrees of freedom that make it hard to track which (out of many possible) gradient trajectory is followed at each point in time. Thus, the same research direction conceives editing mechanisms of few parameters that are applied on a node-by-node basis and reduce the available degrees of freedom to the bare minimum needed to optimize posterior objectives. In this work, we build on this proposition and drastically improve previous heuristics by deriving both which node properties (information available during graph filtering) should be involved in editing and what form editing mechanisms should take. 

To support our analysis, in Subsection~\ref{achieving optimality} we express how tightly prior editing mechanisms should approximate the gradients of graph filter posterior objectives to let the latter reach local optimality with respect to a broad class of twice differentiable objectives. 
In Subsection~\ref{surrogate editing}, we progress our prior editing framework to explain why parameterized (e.g., neural) models with enough degrees of freedom are suited to prior editing. Based on theoretical analysis, in Subsection~\ref{surrogate} we devise appropriate neural networks to be trained at runtime for each prior graph signal and filter so that they estimate ideal priors that trade-off original posterior preservation and meeting fairness constraints. 
After a certain point, our analysis requires that ideal posteriors lie close enough to the original ones outputted by base graph filters. This means that original filters should already procure ``good enough'' first posteriors for downstream objectives. 

\subsection{Local optimality of prior editing}\label{achieving optimality}
In this subsection we investigate theoretical properties of positive definite graph filters that allow coarse approximations of optimal prior editing schemes to reach local optimality with regards to their induced posterior objective values. Before going through our main analysis, we point out that real-world graph filters often end up with a post-processed version of posteriors. For example, personalized pagerank is often implemented as an iterative application of the power method scheme $r= aWr+(1-a)q$ that quickly converges to small numerical tolerance by performing L1 normalization after each step (e.g., as \cite{lin2010power} do), i.e., by dividing node posteriors with their sum. This does not affect the ratio of importance scores placed on propagating prior graph signals different number of hops away, but produces scaled scores that sum to $1$.
\par
More complex post-processing mechanisms may induce different transformations per node that can not be modeled by graph filters. Taking these concerns into account, we hereby introduce a notation with which to formalize post-processing and integrate it in our proposed approach; we consider a \textit{post-processing vector} with which posteriors are multiplied element-by-element. Mathematically, this lets us write post-processed posteriors $r$ of passing graph signal priors $q$ through a graph filter $F(\hat{A})$ as: 
\begin{equation}\label{post-processing}
    r = \text{diag}(p) F(\hat{A}) q
\end{equation}
The exact transformation arising from post-processing mechanisms could vary, depending on both the graph filter and graph signal priors. However, we can decouple this dependency by thinking of the finally selected post-processing as one particular selection out of many possible ones. For instance, L1 output normalization can be  modeled as multiplication with the inverse of the original posteriors' L1 norm $p[v]=\tfrac{1}{\|F(\hat{A})q\|_1}$.

Given a graph filter with post-processing, we now introduce the concept of approximately tracking the optimization slope of posterior objectives with small enough error. This is formalized in Definition~\ref{optimizer}, which introduces a class of multivariate multivalue functions, called $\lambda$-optimizers of objectives, that approximate the negative gradients of objectives with fixed relative error bound $\lambda$. Smaller values of this strictness parameter indicate tighter approximation of optimization slopes, whereas smaller values indicate looser tracking. To disambiguate the possible directions of slopes, our analysis considers loss functions of non-negative values to be minimized. 

\begin{definition}\label{optimizer}
A continuous function $f:\mathcal{R}\to \mathbb{R}^{|\mathcal{V}|}$ will be called a $\lambda$-optimizer of a loss function $\mathcal{L}(r)$ over graph signal domain $\mathcal{R}\subseteq \mathbb{R}^{|\mathcal{V}|}$ only if: 
\begin{equation*}
    \|f\big(r\big)+\nabla \mathcal{L}(r)\|< \lambda \|\nabla \mathcal{L}(r)\|\quad \text{ for all }\nabla \mathcal{L}(r)\neq \textbf{0}
\end{equation*}
\end{definition}

We now analyse the robustness of positive definite graph filters with post-processing in terms of how tight optimizers of posterior losses should be for the graph filter's propagation mechanism to ``absorb'' relative errors and eventually reach local minima. To this end, in Lemma~\ref{robustness} (proof in Appendix~\ref{proofs}) we find a maximum tightness parameter sufficient to lead to local optimality of posteriors with respect to the loss. The required tightness depends on the graph filter's maximum and minimum eigenvalues and the post-processing vector's maximum and minimum values (these depend on both the filter and the normalized adjacency matrix's eigenvalues). For results to hold true, the graph filter needs to be symmetric positive definite. The fact that non-exact optimizers suffice to track trajectories supports the rest of our analysis. 

\begin{lemma}\label{robustness}
Let $F(\hat{A})$ be a positive definite graph filter and $p$ a post-processing vector. If $f(r)$ is a $\tfrac{\lambda_1 \min_v p[v]}{\lambda_{\max} \max_v p[v]}$-optimizer of a differentiable loss $\mathcal{L}(r)$ over graph signal domain $\mathcal{R}\subseteq\mathbb{R}^{|\mathcal{V}|}$, where $\lambda_1,\lambda_{\max}>0$ are the smallest positive and largest eigenvalues of $F(\hat{A})$ respectively, updating graph signals per the rule:
\begin{equation}\label{update}
\begin{split}
    &\frac{\partial q(t)}{\partial t}=f(r(t))
    \\&r(t)=\text{diag}(p)F(\hat{A})q(t)
\end{split}
\end{equation}
asymptotically leads to the loss to local optimality if posterior updates are closed in the domain, i.e. $r(t)\in \mathcal{R}\,\forall t\in[0,T]\Rightarrow \text{diag}(p)F(\hat{A})\int_0^{T}f\big(r(t)\big)dt\in\mathcal{R}$.
\end{lemma}

Following a similar approach as to check whether graph filters are positive definite in Equation~\ref{condition}, we can also bound the eigenvalue ratio $\tfrac{\lambda_1}{\lambda_{\max}}$ involved in calculating the necessary approximation tightness when we know only the type of filter but not the graph or priors. In particular, for filters $F(\hat{A})$ where $\hat{A}$ are symmetrically normalized adjacency matrices of undirected graphs it holds that:
\begin{equation}
    \frac{\lambda_1}{\lambda_{\max}}\geq \frac{\min_{\lambda \in[-1,1]} F(\lambda)}{\max_{\lambda\in[-1,1]}F(\lambda)}
\end{equation}
Such bounds are typically lax. To demonstrate this, we compute them for the personalized pagerank and heat kernel filters, which can respectively be expressed in closed forms as $(1-a)(I-a\hat{A})^{-1}$ and $e^{-t(I-\hat{A})}$, where $a$ and $t$ are their parameters. For these filters, their respective eigenvalue ratios for $\lambda\in[-1,1]$ are at most $\tfrac{1-a}{1+a}$ and $e^{-2t}$. For larger values of $a$ and $t$, which penalize less the spread of graph signal priors farther away, stricter optimizers are required to keep track of the gradient's negative slope. When normalization is the only post-processing employed, all elements of the personalization vector are the same and bounds for sufficient optimizer strictness coincide with the aforementioned eigenvalue ratios. Inserting some specific numbers, for the most widespread version of personalized pagerank with $a=0.85$ it suffices to select $0.081$-optimizers to edit priors. Even for wider prior diffusion with $a=0.99$ it suffices to select $0.005$-optimizers. When graphs have adequately many nodes, such optimizers are orders of magnitude larger than the average posteriors $\tfrac{1}{|\mathcal{V}|}$ arising from L1 normalization. 

In Subsection~\ref{threats to applicability}, we explain why larger eigenvalue ratios broaden the applicability of our mathematical analysis and identify threats that arise when those filters that focus on many propagations---and thus require tight optimizers---are applied on graphs with too few nodes. In that scenario, optimization errors become comparable to the average posteriors and therefore generate untraceable optimization trajectories.


\subsection{Surrogate models for prior signal editing}\label{surrogate editing}
The above analysis lets us express what objectives like fair graph filtering would look like under the prism of accrediting their quality (e.g., biases) to priors. In particular, we try to meet posterior objectives by employing appropriate $\lambda$-optimizers of objective gradients to adjust priors per Equation~\ref{update}. To this end, we argue that finding the editing mechanisms and computing line integrals leading to locally optimal priors is unnecessary; instead, we can create models that directly compute the integration outcome.

This intuition holds on a theoretical level too; Lemma~\ref{anyedit} (proof in Appendix~\ref{proofs}) transcribes the required tightness of optimizers around loss gradients to near-invertibility of a prior editing model's Jacobian across the gradients' direction. If the requirement is met, there exist prior editing model parameters that lead posteriors to locally optimize the objective. Since the same quantity as in Lemma~\ref{robustness} identifies adequate tightness, even coarse prior editing models can be met with success.

\begin{lemma}\label{anyedit}
Let us consider graph signal priors $q_0$, positive definite graph filter $F(\hat{A})$ with largest and smallest eigenvalues $\lambda_{\max},\lambda_1$, post-processing vector $p$, differentiable loss function $\mathcal{L}(r)$ with at least one minimum in domain $\mathcal{R}\subseteq\mathbb{R}^{|\mathcal{V}|}$, and a differentiable graph signal generation function $\mathcal{M}:\mathbb{R}^K\to\mathbb{R}^{|\mathcal{M}|}$ for which there exist parameters $\theta_0$ satisfying $\mathcal{M}(\theta_0)=q_0$ and $\text{diag}(p)F(\hat{A})\mathcal{M}(\theta)\in\mathcal{R}$. If, for any parameters $\theta$, its Jacobian $\mathbb{J}_\mathcal{M}(\theta)$ has linearly independent rows and satisfies:
\begin{align*}
    &\big\|E(\theta)\nabla \mathcal{L}(r)\big\|< \tfrac{\lambda_1 \min_v p[v]}{\lambda_{\max} \max_v p[v]}\|\nabla \mathcal{L}(r)\|\quad\text{ for }\nabla \mathcal{L}(r)\neq \textbf{0}\\
    &E(\theta) = \mathcal{I}-\mathbb{J}_\mathcal{M}(\theta)(\mathbb{J}_\mathcal{M}^T(\theta)\mathbb{J}_\mathcal{M}(\theta))^{-1}\mathbb{J}_\mathcal{M}^T(\theta)\\
    &r(\theta) = \text{diag}(p)F(\hat{A})\mathcal{M}(\theta)
\end{align*}
then there exist parameters $\theta_{\infty}$ that make $\mathcal{L}\big(r(\theta_\infty)\big)$ locally optimal.
\end{lemma}

\par
In Lemma~\ref{anyedit}, the matrix $E$ computes differences between the unit matrix and multiplying the Jacobian matrix  $\mathbb{J}_\mathcal{M}(\theta)$ with its right pseudo-inverse; differences are constrained to matter only for nodes with high gradient score values. Thus, as long as the objective's gradient retains a clear---though maybe changing---direction to move towards to and this can be captured by a surrogate prior editing model $\mathcal{M}(\theta)$, which may (or should) fail at following other kinds of trajectories, then a parameter trajectory path exists to arrive at locally optimal prior edits.
For example, if prior editing had the same number of parameters as the number of nodes and its parameter gradients were linearly independent, $\mathbb{J}_\mathcal{M}(\theta)$ would be a square invertible matrix, yielding $\mathbb{J}_\mathcal{M}(\theta)(\mathbb{J}_\mathcal{M}^T(\theta)\mathbb{J}_\mathcal{M}(\theta))^{-1}\mathbb{J}_\mathcal{M}^T(\theta)=\mathcal{I}\Leftrightarrow E=\textbf{0}$ and the theorem's precondition inequality would always hold. Whereas, as the number of parameters decreases, it becomes more important for $\mathcal{M}(\theta)$ to be able to exhibit degrees of freedom in the same directions as its induced loss's gradients. Next, we aim to learn the directions of the degrees of freedom with deep neural models.

\subsection{Neural approximation of posterior optimization with prior signal editing}\label{surrogate}
From a high-level standpoint, Lemma~\ref{anyedit} suggests that, for each prior graph signal $q_0$ and a graph filter posterior objective, there exists 
an appropriate prior editing mechanism $\mathcal{M}(\theta)$ that follows a differentiable trajectory towards a point of locally optimal priors $\mathcal{M}(\theta_\infty)$ corresponding to parameter values $\theta_\infty$. Found priors are considered locally optimal both in the sense that they end up locally optimizing the objective and in that they need to lie close enough to the original ones to be discovered. 

We now propose that neural network architectures are valid prior editing mechanisms; thanks to the universal approximation theorem (we use the form presented by \cite{kidger2020universal} to work with non-compact domains and obtain sufficient layer widths), they can produce tight enough approximations of the desired objectives to serve as $\lambda$-optimizers. The minimum requirements neural architectures need to satisfy are summarized in Theorem~\ref{mlp} (proof in Appendix~\ref{proofs}). Importantly, the theorem specifies appropriate neural architectures and several hyperparameters (e.g., their width), but their ideal depth can only be acquired via hyperparameter tuning on each filtering task.

\begin{theorem}\label{mlp}
Let us consider graph signal priors $q_0$, positive definite graph filter $F(\hat{A})$ with largest and smallest eigenvalues $\lambda_{\max},\lambda_1$ respectively, post-processing vector $p$, and 
twice differentiable loss function $\mathcal{L}(r)$, with some unique minimum at unknown posteriors $r=r_\infty$ of a domain $\mathcal{R}\subseteq\mathbb{R}^{|\mathcal{V}|}$, and whose Hessian matrix $\mathbb{H}_\mathcal{L}(r)$ linearly approximates its gradients $\nabla\mathcal{L}(r)$ within the domain with error at most $\epsilon_\mathbb{H}$. Let us also construct a node feature matrix $H^{(0)}$ whose columns are graph signals (its rows $H^{(0)}[v]$ correspond to nodes $v$). If $q_0$ and all graph signals involved in the calculation of $\mathcal{L}(r)$ are columns of $H^{(0)}$, and it holds that: $$\tfrac{\lambda_{1} \min_v p[v]}{\lambda_{\max} \max_v p[v]}\|\nabla\mathcal{L}(r)\|>2\epsilon_\mathbb{H}\|r-r_\infty\|$$
for any $r\in\mathcal{R}\setminus\{r_\infty\}$, then for any $\epsilon_\infty>0$ there exists a deep neural network architecture $$H^{(\ell)}=\phi_\ell(H^{(\ell-1)}W^{(\ell)}+b^{(\ell)})$$ with activation functions 
$$\phi_\ell(x)=\{x\text{ for }\ell=L,relu(x)\text{ otherwise}\}$$ learnable weights $W^{(\ell)}$, and biases $b^{(\ell)}$ for layers $\ell=1,2,\dots,L$ with depth $L\geq 4$, in which using the output of the last layer $H^{(L)}$ as priors yields posteriors arbitrarily close to the local minimum $\|diag(p)F(\hat{A})H^{(L)}-r_\infty\|<\epsilon_\infty$. Additionally, layers other than the last can have output columns equal to the number of $\text{columns of }H^{(0)}+2$.
\end{theorem}

\begin{figure}[htbp]
\centering
    \includegraphics[width=0.7\textwidth,clip]{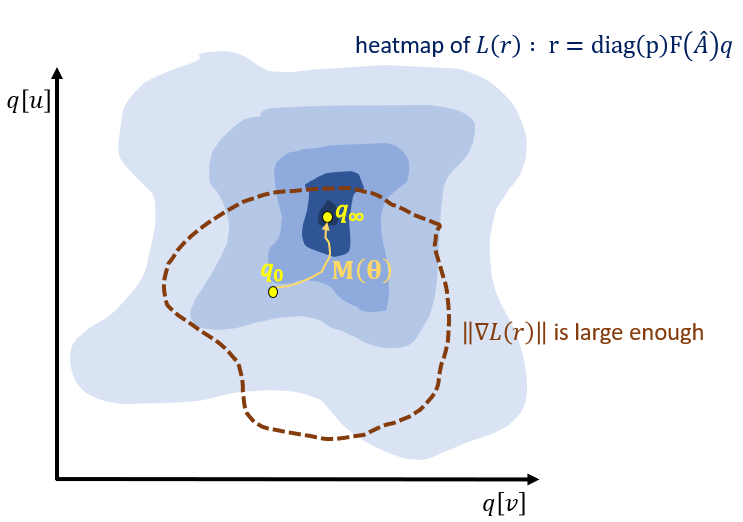}
    \caption{Finding nearby priors that locally optimize graph filtering objectives.}
    \label{fig:attractiveness}
\end{figure}

To summarize the main result of Theorem~\ref{mlp}, there exists a small enough area $\mathcal{R}$ around locally optimal priors $q_\infty$, in which biased posteriors corresponding to priors $q_0$ can still be usable as a reference to approximate $q_\infty$. Such an area is enclosed within the brown dashed curve in Figure~\ref{fig:attractiveness} and is wider the closer to linearity posterior objectives are or the looser $\lambda$-optimizers of graph filters are. At the same time, loss derivatives should be large enough to push optimization trajectories towards minima, even from faraway points of the domain.

\section{Surrogate Graph Neural Networks of Fairness-Aware Graph Filtering}\label{surrogate posteriors}
\begin{figure}[htbp]
\centering
    \includegraphics[width=0.95\textwidth,clip]{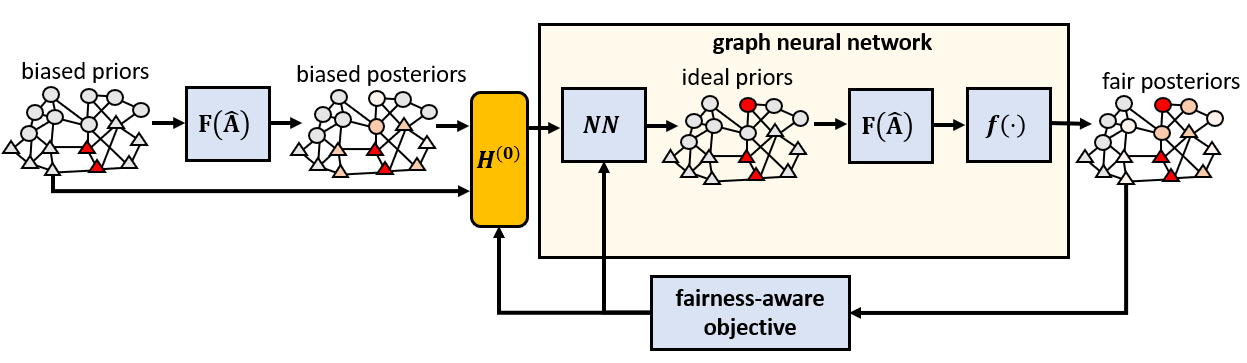}
    \caption{Overview of the proposed fair posterior framework.}
    \label{fig:overview}
\end{figure}

At this point, we have all the necessary tools to tackle the main goal of this work, i.e., to procure new posteriors that are similar to those of graph filters but which also achieve high prule fairness. In this regard, we introduce fairness-aware posterior objectives to be optimized with neural prior editing. At first, we formulate the objectives to trade-off node score preservation and fairness, but later evolve our pipeline to optimize only preservation under hard-coded fairness constraints. Our approach is summarized in Figure~\ref{fig:overview}; starting from only from one prior value for each node, we obtain features $H_0[v]$ for nodes $v$ to input in the neural network \textit{NN}, and train the latter to predict new node priors that in the end lead to fair posteriors. The final step of producing fair posteriors also passes the filter's outcome through a transformation function $f(\cdot)$ that ensures theoretical properties of our analysis transfer to the practical setting at hand (e.g., let guarantees of symmetric normalization apply column-wise normalization). 

The pipeline of learning new priors, passing them through the graph filter, and performing output transformations is equivalent to training a predict-then-propagate graph neural network whose propagation mechanism is the filter itself and the final activation applies postp-processing (e.g., normalization) and transformation functions. In other words, the proposed framework is equivalent to a surrogate graph neural network model of ideal posteriors (Subsection~\ref{architecture}). The same framework and corresponding analysis is in principle applicable to a variety of posterior objectives, though in this work we only operate it for disparate impact mitigation.

In this section, we explore the architecture and hyperparameters of the above framework. First, Subsection~\ref{objective} specifies fairness-aware objectives, and introduces a methodology for adapting them or any other twice differentiable loss to our mathematical analysis by adding an L1 posterior regularization term. Then, Subsection~\ref{features} gathers node features $H^{(0)}$ requisite for Theorem~\ref{mlp} to optimize the desired objective. Subsection~\ref{architecture} presents the exact graph neural network architecture of our approach; 
architecture details and some hyperparameters are to be selected anew for each graph filtering task (e.g., with grid search among promising values), such as when different priors are analysed. Finally, Subsections~\ref{extending} and~\ref{workaround} present activation functions that respectively find local optima for graph filters applied on asymmetric adjacency matrix normalization or objectives prone to numerical instability, and impose fairness constraints without needing to search for appropriate weight trade-offs within fairness-aware objectives.

\subsection{Defining fairness-aware objectives in our framework}\label{objective}
Training prior editing mechanisms so that they preserve the outcome of graph filtering but also become fairness-aware requires appropriate selection of objective/loss functions that satisfy the conditions of Theorem~\ref{mlp}. We devise losses that trade-off disparate impact mitigation and original posterior score preservation:
\begin{equation*}
    \mathcal{L}(r)=\mathcal{L}_{ret}(r) + l_{bias}\mathcal{L}_{bias}(r)
\end{equation*}
where the component $\mathcal{L}_{ret}(r)$ captures the degree of preservation of original posteriors and $\mathcal{L}_{bias}(r)$ quantifies posterior bias. 

Within such losses, $l_{bias}\geq 0$ is a trade-off hyperparameter; larger values indicate more importance being placed on bias mitigation than retaining posteriors. The ideal trade-off is not necessarily fixed, but should assume the smallest possible value needed to reach desired fairness constraints. Too large values may make fairness dominate neural learning processes and thus end up with shallow approximations of original posteriors, but too small values may also fail to induce fairness. Subsection~\ref{workaround} presents a methodology that substitutes the necessary hyperparameter exploration by setting $l_{bias}=0$ and imposing additional post-processing (by modifying the graph neural network's output activation) to achieve perfect disparate impact mitigation. Here we define a full objective for the sake of completeness.

Before fleshing out the loss, we address the concept of universal approximation via enhancing \textit{any} twice differentiable loss to satisfy the preconditions of Theorem~\ref{mlp}. Double differentiability is easy to satisfy, at least within the optimization domains of machine learning frameworks constrained to avoid computational singularities; these domains are often implicitly defined in machine learning tasks, given that optimization trajectories tend to avoid singularities in probability and thus there could exist equivalents that match the same points functions are evaluated at but also bypass the singularities.
However, the theorem also requires that gradient magnitudes should be significantly larger compared to the error of linearly approximating them via the Hessian, so as to overcome local optima. To facilitate universal applicability of our framework, i.e., to losses that may not necessarily satisfy the last property, we introduce the practice of \textbf{adding sufficiently large L1 posterior regularization} to base losses $\mathcal{L}(r)$, resulting to the following regularized versions:

\begin{equation}\label{loss adjustment}
    \tilde{\mathcal{L}}(r) = \mathcal{L}(r)+{l_{reg}}\int_{\mathcal{M}(r)} sgn(r)\cdot dr \approx \mathcal{L}(r)+{l_{reg}}(\|r\|_1-\|r_0\|_1)
\end{equation}
where $\cdot$ is the dot product, $sgn(\cdot)$ is a twice differentiable tight enough approximation of the element-by-element sign operator, $l_{reg}\geq 0$ a sufficiently large regularization parameter, and $\mathcal{M}(r)\subseteq\mathcal{R}$ is an (unknown) optimization trajectory within domain $\mathcal{R}$ that starts from $r_0$ and asymptotically approaches $r$. Theorem~\ref{adjust theorem} (proof in Appendix~\ref{proofs}) explains why large enough L1 regularization introduces a constant push of posteriors towards zero that keeps gradients sufficiently large. The upper bound we provide for the regularization is a sufficient condition. Recall that the optimization domain can attract many but not all possible posteriors to locally optimal priors. Hence, graph filters employed in practice should be selected to form close enough approximations of the desired objective (i.e., not all filters are suited to all graphs and objectives).

\begin{theorem}\label{adjust theorem}
    For any twice differentiable loss $\mathcal{L}(r)$, there exists sufficiently large enough parameter $l_{reg}\in\big[0,2\tfrac{\sup_q\|q-q_0\|}{|\mathcal{V}|}\big]$ such that the loss regularization of Equation~\ref{loss adjustment} satisfies the properties needed by Theorem~\ref{mlp} within the optimization domain $$\mathcal{R}=\{r_\infty\}\cup\bigg\{r:\|r-r_\infty\|<0.5\max\big\{|\mathcal{V}|,\tfrac{\|\nabla\mathcal{L}(r)\|}{\epsilon_\mathbb{H}}\big\}\tfrac{\lambda_{1} \min_v p[v]}{\lambda_{\max} \max_v p[v]}\bigg\}$$where $r_{\infty}$ is the ideal posteriors optimizing the regularized loss. If the second term of the $\max$ is selected, it suffices to have no regularization $l_{reg}=0$.
\end{theorem}

Given the freedom in loss selection granted by the regularization term, we now focus solely on the business objective of surrogate graph filtering that addresses the disparate impact of node scores. First, we aim to minimize the utility loss of Subsection~\ref{fairness background}) by setting $\mathcal{L}_{ret}(r)=\mathcal{L}_{util}$. Second, we quantify disparate impact mitigation via the prule; to reach perfect mitigation, this measure needs to compute to $1$ and its penalization can be directly used as a loss component, i.e., $\mathcal{L}_{bias}(r)=1-prule$. 

\subsection{Node features}\label{features}
Overall, when applying Theorem~\ref{mlp} to fairness-aware graph filtering, it suffices to introduce only three node feature dimensions for the graph neural network to process. First, original priors need to be inputs of the architecture. This enables the perfect replication of original posteriors requisite for Lemma~\ref{anyedit}, for instance when the neural prior editing mechanism just outputs the original priors. 
Second, original posteriors are directly used in the loss function calculation. Third, the sensitive attribute indicates which nodes belong to the protected group and is also a type of graph signal (a value that depends on the node) contributing to the computation of the loss. 

All three signals should be set as node feature dimensions. Thus, we extract the following node features $H^{(0)}:|\mathcal{V}|\times 3$ to input in graph neural network architectures:
\begin{align*}
&H^{(0)}[v,0]=q_0[v]
\\&H^{(0)}[v,1]=r_0[v]
\\&H^{(0)}[v,2]=\{1\text{ if }v\in {S},0\text{ otherwise}\}
\end{align*}
where, $q_0$ are the original priors, $r_0=diag(p)H(\hat{A})q_0$ the original posteriors for graph filter $H(\hat{A})$ with post-processing vector $p$, and $S$ is the group of protected nodes. No other signals contribute to the computation of the previous subsection's loss, which means that $H^{(0)}$ suffice to generate near-ideal priors. This analysis holds true for symmetric normalized graph adjacency matrices and is extended to asymmetric ones in Subsection~\ref{extending} by adding a fourth node feature dimension. 

\subsection{Architecture}\label{architecture}

Given a loss function and node feature matrix $H^{(0)}$, both of which are appropriately constructed as in the previous subsections to meet the requirements of Theorem~\ref{mlp}, we formally express the theorem's estimation of ideal posteriors organized into a column vector $H^{(L)}$ via a model of the following form:
\begin{align*}
&H^{(\ell)}=\text{relu}(H^{(\ell-1)}W^{(\ell)}+b^{(\ell)})\quad\ell=1,2,\dots,L\\
&\hat{r}_\infty=diag(p)F(\hat{A})H^{(L)}
\end{align*}
for post-processing vector $p$, and appropriate layer weights and biases $W^{(\ell)}$ and $b^{(\ell)}$ respectively. This formula resembles the predict-then-propagate architecture of Subsection~\ref{gnns} and can therefore be implemented in graph neural network frameworks supporting related operations. But, contrary to the widespread practice of engaging the same predefined heuristic architectures each time, our analysis reveals several necessary choices listed below.

First, we employ neural layers of $3+2=5$ dimensions---or $4+2=6$ dimensions for asymmetric adjacency matrix normalization. This is different than the larger number (e.g., $64$ or $128$) of hidden layer dimensions in traditional graph neural networks, but is justified in that we are not working with too many node features and thus require a smaller number of latent dimensions. Similarly to the predict-then-propagate scheme, we employ \textit{relu} activations at intermediate layers. However, at the top layer, outputs are not used for classification; given that we need the architecture to be able to perfectly reconstruct base graph filtering posteriors despite not explicitly using those as inputs, the output activation approximating this concept should ideally remain linear. Thus, the only transformation we apply is the post-processing vector multiplication already employed by filters and the filter transfer trick, which is a complementary mechanism that will be developed in Subsection~\ref{extending}.

Second, the graph filter performing the propagation of neural logits $H^{(L)}$, where in our case these predictions are the edited priors, should be identical to the one producing the original potentially biased posteriors. For instance, if iterative formulas are used to compute graph filters until numerical tolerance  (e.g., the power iteration for personalized pagerank), the same number of iterations need to be repeated during propagation, regardless of the latter's convergence properties. From a theoretical standpoint, breaking the premise of reusing the same kind of filtering with a different number of iterations would make underlying optimization trajectories non-continuous and thus makes applicability of our analysis uncertain. The negative effect of using different types of filtering together is demonstrated in the ablation study of Appendix~\ref{ablation study}. 

A third and final consideration, which plays a vital role in the generalization abilities of neural networks but has not been addressed so far, is how neural parameters should be initialized before training starts. To avoid shallow minima for deep graph neural networks, \cite{glorot2010understanding} recommend parameter initialization strategies that assign randomized values to dense matrix transformations while preserving the expected variance of inputs and therefore prevent vanishing or exploding gradients at the first optimization steps. For \textit{relu} activations, \cite{he2015delving} show that this property is satisfied at layer $\ell$ for zero initial biases and dense weights sampled from a normal distribution $\mathcal{N}\big(0,\sigma^{(\ell)}\big)$ with zero mean and standard deviation $\sigma^{(\ell)}=\sqrt{\tfrac{2}{\text{columns of }W^{(\ell)}}}$. We devote the rest of this subsection to adjusting initialization so that it avoids pitfalls stemming from narrow neural layer widths.

When initialization weights are sampled from a distribution that is symmetric around zero, and which has therefore zero mean, narrow layer widths like ours tend to create a lot of ``dying'' neurons, where \textit{relu} does not backpropagate during training due to activations that reside in its non-positive region \citep{lu2019dying}. This poses a significant risk in the ability of training our proposed neural architecture; at worst, for some nodes its single output may not contribute to training. This concern is addressed in the literature either by adjusting activation functions to produce small but non-zero derivatives for negative inputs \citep{he2015delving}, or by employing asymmetric initialization strategies \citep{lu2019dying}. We choose the last type of approach, because our analysis is applied on potentially non-compact domains (e.g., where objectives are not differentiable at certain points); to work on such domains we stick to the universal approximation theory results of \cite{kidger2020universal}, who only present guarantees for \textit{relu} under non-compactness. 

Since propagating graph signals with non-negative values yields non-negative posteriors, we elect to use a folded normal distribution for initialization, i.e., which applies an absolute value on the normal distribution's samples and thus preserves the positive sign of all inputs for the posteriors arising from initialization. Given that we start from a normal distribution with zero mean, the standard deviation of the folded variation \citep{tsagris2014folded} becomes $\phi_{fold}=\sigma\sqrt{1-\tfrac{2}{\pi}}$, where $\sigma$ is the standard deviation of the zero-mean normal distribution on which absolute value folding is applied. Thus, to preserve the variance of inputs, we initialize weights per:
\begin{equation}
    W^{(\ell)}=\big|W_{normal}^{(\ell)}\big|: W_{normal}^{(\ell)}\sim\mathcal{N}\bigg(0, \sqrt{\frac{2}{(1-\tfrac{2}{\pi})\text{ columns of }W^{(\ell)}}}\bigg)
\end{equation}

\subsection{The filter transfer trick}\label{extending}
The theoretical groundwork and graph neural network architecture we presented so far are only applicable to filters of symmetrically normalized adjacency matrices. However, other types of normalization also hold practical value. For example, the adjacency matrices of directed graphs are typically asymmetric, whereas personalized pagerank filters often model Markov chains via column-wise normalization. Additionally, objectives involving division with small original posteriors, such as the relative errors in the unbiased utility loss of Equation~\ref{utility}, tend to make low-scored node errors dominate the optimization process. In this subsection, we devise a blanket methodology that addresses both of these issues, which we dub the \textit{filter transfer trick}. This extends our approach to both asymmetric filtering, i.e., the outcome of filtering asymmetric adjacency matrices, and numerically unstable objectives by devising appropriate twice differentiable transformation functions $f:\mathbb{R}^{|\mathcal{V}|}\to\mathbb{R}^{|\mathcal{V}|}$ to generate posteriors $f(r)$ in ``problematic'' settings from the posteriors $r$ learned for symmetric graph filtering on numerically stable objectives.

We first express the filter transfer trick for asymmetric filtering. In this setting, we search for posteriors $r_{ns}=diag(p_{ns})F(\hat{\mathcal{A}}_{ns})q$ that locally minimize losses $\mathcal{L}_{ns}(r_{ns})$ for asymmetrically normalized adjacency matrices $\hat{\mathcal{A}}_{ns}$ and post-processing vectors $p_{ns}$ with non-zero elements. Thus, to reproduce asymmetric filtering in an optimize-able manner, we depend on posteriors $r=diag(p)F(\hat{\mathcal{A}})q$ obtained for symmetrically normalized adjacency matrices $\hat{\mathcal{A}}$ and any post-processing $p$ with non-zero elements, and compute the transformations $r_{ns}=f(r)$ for which $\mathcal{L}_{nr}\big(f(r)\big)$ is minimized. Given that appropriate transformations exist, it suffices to find $r$ with our graph neural network methodology to minimize losses $\mathcal{L}(r)=\mathcal{L}_{ns}\big(f(r)\big)$. 

A barrier in applying the above methodology is that, to avoid shallow minima, transformation functions should also preserve structural information captured by asymmetric filtering. Future research can consider learning the functions, for instance via adversarial techniques, but this requires extensive analysis and experimentation that lie outside the scope of this work. Instead, we conceive functions that, for many common settings, find locally optimal posteriors $r_{ns}$ of asymmetric adjacency matrix filtering that lie close to original posteriors $r_{ns0}$ of unedited priors $q_0$, and exhibit the same receptive breadth as original filters, i.e., let nodes be influenced from the same number of hops away.

Finding posteriors close to original ones is similar to what we achieved for symmetric filtering, but this time we do not maintain all implicit structural characteristics of asymmetric filtering (the receptive breadth is only one characteristic) and thus run the risk of discovering shallow optima. Nonetheless, we make the assumption that too shallow optima are avoided thanks to different kinds of adjacency normalization following roughly the same structural analysis principles of graph filters. Given the mathematical intractability of working with abstract types of normalization, we only experimentally validate this claim for column-wise normalization.

The transformation functions we study come from the following family:
\begin{equation}\label{transform}
f_{\delta}(r)=r_{ns0}\tfrac{\delta+r}{\delta+r_{0}}
\end{equation} where all operations are applied element-by-element on graph signals, $r_{ns0}$ are the original asymmetric filter posteriors, and $r_{0}$ are the original symmetric filter posteriors arising from the unedited priors $q_0$. The parameter $\delta\geq 0$ trades-off numerical stability for larger values vs. fast learning for smaller values. In particular, given that the loss has a bounded derivative (i.e., is Lipschitz continuous), it holds that  $\lim_{{\delta}\to\infty}\nabla\mathcal{L}(f_\delta(r))=\textbf{0}$. During hyperparameter selection, we search among potential values $\delta=\delta_0\max_{v}r_0[v]$ where $\delta_0\in\{0.1,1,10\}$. These values control optimization robustness (Subsection~\ref{sec:robustness}) while remaining comparable to the order of magnitude of filtering outcomes, thus providing leeway for derivative-based optimization.

In Theorem~\ref{trick opt} (proof in Appendix~\ref{proofs}) we show that, for the class of filters with non-negative coefficients, non-negative prior node values, and adjacency matrix normalization with positive edge weights (e.g., column-wise normalization), the proposed transformation can find locally optimal posteriors for asymmetric filtering based on a dual symmetric problem. Additionally, the theorem defines an equally hard to solve symmetric problem, which determines how tight neural approximation of the symmetric filter should be.

\begin{theorem}\label{trick opt}
    Let us consider the case where priors and filter parameters are all non-negative, and asymmetric adjacency matrix normalization yields positive edge weights. Equation~\ref{transform} can express locally optimal asymmetric filtering posteriors, which can be found by applying the graph neural architecture of this section on the equivalent symmetric filter to optimize the loss $\mathcal{L}(f_{\delta}(r))$ from node features $H^{(0)}$ extended with a fourth dimension $H^{(0)}[v,3]=r_{ns0}[v]$ and $6$ neural layer dimensions. The optimization tightness of the symmetric filter is the same as if we minimized $\mathcal{L}\big(diag\big(\tfrac{p}{\delta+r_0}\big)F(\hat{\mathcal{A}})q\big)$.
\end{theorem}

The proof of the above theorem can be generalized to locally optimize posteriors of any filter with a predefined symmetric filter of the same receptive breadth instead of only the symmetric counterpart of the asymmetric filter. Although it is tempting to find local optima this way, when different filters are used to optimize each other's posteriors there is a non-negligible risk of reaching shallow minima (Appendix~\ref{ablation study}), similar to the concerns for unconstrained posterior optimization. A more thorough discussion of this point is reserved for Subsection~\ref{threats to applicability}. For now, we stress that, when applied to asymmetric filtering, this subsection's approach is a heuristic tailored to filters of the same functional forms. 

The family of transformations presented in Equation~\ref{transform} is also applicable on symmetric filtering to guide optimization towards posteriors with the same order of magnitude as original ones. In this case, the transformed posteriors $f(r)$ are tied to the same filter form, adjacency matrix $\hat{A}_{ns}=\hat{A}$, and original posteriors $r_{ns0}=r_0$ as the the ones used to compute $r$. Thus, the transformation collapses to $$f(r)=r_0\tfrac{\delta+r}{\delta+r_0}$$ and maintains near-identical propagation mechanisms to the original filter as $\delta\to 0$. The main difference becomes that gradients remain proportional to the order of magnitude of $r_0$'s elements, which prevents optimizers from abruptly inducing large score changes to low-scored nodes. For this case, we entertain the same hyperparameter search for values $\delta$ as before. Furthermore, the transformation function does not inject any new graph signal in the computation of the loss $\mathcal{L}(f(r))$, which means that no additional columns should be concatenated to the node feature matrix $H^{(0)}$, i.e., the original columns can be maintained when making symmetrc filtering fairness-aware.

As a final remark, although the filter transfer trick can port our approach to some problematic settings and even control the optimization robustness, the family of transformation functions we employ does not naturally extend to negative priors. 


\subsection{Hard-coded fairness constraints}\label{workaround}
When fairness is integrated into the loss component of our architecture, as we previously did, it requires careful hyperparameter exploration to achieve fairness while avoiding getting stuck at unfair yet Pareto optimal posteriors, for example due to different scale between the prule and utility loss derivatives. However, the exploration can be computationally costly and therefore limit the practical viability of our approach (Subsection~\ref{threats to validity}). To address this issue, this subsection presents another application of the filter transfer trick's principles to algorithmically impose fairness constraints instead of incorporating them in the loss; we rely on posterior transformation functions that make their outputs always meet desired fairness constraints and remove respective terms from the desired loss. 

To derive such functions in the context of mitigating disparate impact, which is the focus of this work, we look at the simple methodology for balancing non-negative posterior mass between protected and non-protected groups $S$ and $\mathcal{V}\setminus S$ by upscaling the node scores of the one and downscaling the scores of the other. For binary sensitive attribute $s[u]=\{1\text{ if }u\in S, 0\text{ otherwise}\}$, the appropriate scaling is achieved with the formula:
$$f_{mult}(r)[v]=p_0\bigg(\frac{(1-\phi)\,s[v]}{\sum_{u\in S}s[u]r[u]}+\frac{\phi\,(1-s[v])}{\sum_{u\not\in S}s[u]r[u]}\bigg)r[v]$$
where $p_0$ is a normalization term to ensure that the same type of scalar multiplication (if any) is imposed on $r_{mult}$ as was used to procure $r$. The balancing process hard-codes $\phi$-fairness adherence (recall that this is equivalent to specific prule values, i.e., $prule=1$ for $\phi=\frac{|S|}{|\mathcal{V}|}$) in our setting, and is applied element-by-element on posteriors $r$. It can also be generalized to multi-value sensitive attributes by diving each group of nodes with the same values with their sum, but we do not experiment with such scenarios in this work. 

The transformation function $f_{mult}(r)$ does not inject any new graph signal in the computation of filter transfer loss $\mathcal{L}(f_{mult}(r))$, which means that no additional columns should be concatenated to the node feature matrix $H^{(0)}$. Furthermore, the practice remains compatible with our analysis. Thus, now that a fairness constraint is enforced by applying an appropriate transformation as part of our architecture's output activation, we can go back to minimizing the utility loss---plus any necessary regularization term---with our framework. Devising a fairness-imposing transformation was simple for disparate impact mitigation, but it may be more complicated for other types of fairness encountered in future work.

\begin{figure}[htbp]
\centering
    \includegraphics[width=0.7\textwidth,clip]{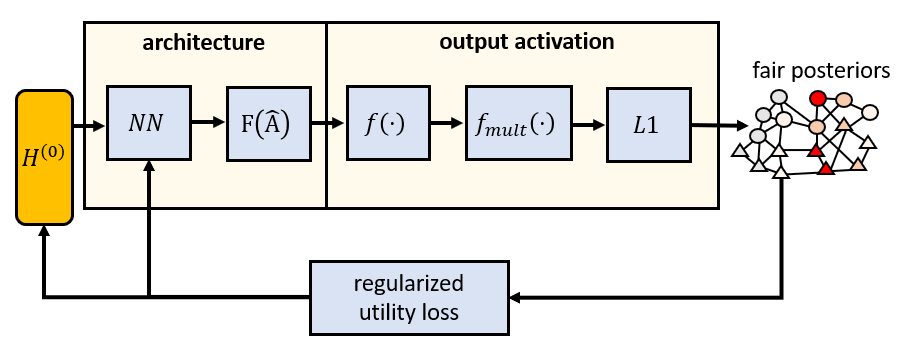}
    \caption{Detailed view of the proposed NSGFF system.}
    \label{fig:architecture}
\end{figure}

Figure~\ref{fig:architecture} presents the graph neural network that integrates into output activations the filter transfer trick $f$, fairness constraint imposition $f_{mult}$, and L1 normalization of final posteriors to be compatible with the normalization employed by graph filters we experiment on. Other types of normalization could also be used. The figure captures implementation details that were not visible in the introductory overview of Figure~\ref{fig:overview}. We refer to this system as neural surrogate graph filter fairness (NSGFF).

\section{Experiments}
In this section we assess whether prior editing can mitigate disparate impact while largely preserving the quality of original posteriors in graph mining tasks. First, in Subsection~\ref{graphs} we describe benchmark filtering tasks; these comprise different graphs and prior graph signals, and define quantitative assessment of filter posteriors. In Subsection~\ref{compared approaches} we present competing approaches for injecting fairness in graph filters. The methodology we follow to explore their efficacy is presented in Subsection~\ref{methodology}. Finally, in Subsection~\ref{results} we run experiments and present results.

\subsection{Graph mining tasks}\label{graphs}
Our exploration spans the two types of downstream graph filtering tasks presented in Subsection~\ref{community}; community member recommendation and graph diffusion. In both tasks, we account for a sensitive graph node attribute, and aim to fully mitigate its disparate impact on filter posteriors by achieving prule $=1$. The two types of mining tasks are each replicated under three ways of emulating how real-world priors would be constructed (see below), and under both fairness constraints. In total, we investigate $2\cdot 3=6$ task variations.

Community member recommendation experiments are conducted on publicly available graphs that comprise metadata communities (e.g., node classification labels or known overlapping communities) and a binary sensitive attribute to be protected from a business, ethical, or legal standpoint. Our work is also applicable to other types of fairness, multi-value sensitive attributes, and constraints that exclude certain nodes from fairness evaluation. However, as other literature approaches do not always account for such settings, we experiment on disparate impact mitigation of a binary sensitive attribute as a common ground. To assess member recommendation, we randomly use either 10\%, 30\%, or 50\% of known community members to construct binary graph signals, and use other known members as test data, for which aim to assign higher posterior scores than non-community members. 
This assessment is quantified with AUC. We experiment on the following three graphs: 

\begin{itemize}
    \item[] \textit{Citeseer \citep{sen2008collective}.} A network of scientific publications among six scientific topics that are linked based on citations. We consider the first two of these topics as the communities whose members we aim to rediscover, for instance as if they were new keywords whose entries we needed to populate. In line with many fairness benchmarks \citep{dong2022fairness}, we protect recommendation results with respect one of the topics---in our case, the one with the most members---so that they obtain on average equal posteriors to the rest of recommended nodes.

    \item[] \textit{Highschool \citep{mastrandrea2015contact}.} A network of highschool student interactions. From this dataset, we extract known friendship relations in Facebook, the classes students attend, and student gender (male/female/unknown). We perform community recommendation experiments, in which we aim to recommend classes to students that are structurally proximate to others already attending the class. During this process, we protect female students to not be assigned on average lower recommendation scores while striving to maintain the original AUC for male students.


    \item[] \textit{Pokec \citep{takac2012data}.} A social network dataset of the namesake platform, where user profiles are linked based on friendship relations. To reduce the running time of experiments, we construct a subgraph with the first 100,000 (out of more than 30 million) edges found in the original dataset. Using these, we aim to recommend users for the cooking and sports communities. At the same time, we consider the binary gender attribute of user profiles (male or female) to be sensitive in the sense that recommendation scores should achieve average parity between the respective groups of users.
\end{itemize}

We also experiment with diffusing prior scores through graphs whose nodes have sensitive labels. For this task, we either diffuse priors for which a fraction of node values among 30\%,50\%,70\% are non-zero and uniformly sampled from the range $[0,1]$. We aim to mitigate disparate impact when diffusing such priors while in large part minimizing the utility loss compared to original (biased) graph filtering. 
In addition to the previous graphs involved in community member recommendation experiments, diffusion is also tested on the following two publicly available graphs:

\begin{itemize}
\item[] \textit{Polblogs.} A network of blogs mined in 2005 that form edges based on the hyperlinks between them. We consider the political opinions expressed in each blog to be a sensitive attribute that should not influence data mining outcomes. For example, this concern may arise when trying to cover news stories, where understanding both sides of an opinion hinters the creation of political echo chambers, and lets stakeholders (e.g., journalists) glean a spherical understanding by not suppressing a specific type of opinion. Experiments of starting from a few nodes replicate a real-world setting of searching for blogs related to a given topic.

\item[] \textit{Polbooks.} A network of political science books that are linked based on whether they have been frequently co-purchased. The books are marked based on the political opinion of left, right, or neural. To convert these values into a binary sensitive attribute, we consider the protection of left vs non-left political opinions.


\end{itemize}

We treat the edges of all the above graphs as undirected. The number of nodes and edges, alongside the number of experimented communities, the number of nodes in the largest community, and the number of protected group members (i.e., the nodes with the protected sensitive attribute value) are summarized in Table~\ref{tab:data sets}. 
\begin{table}[htbp]
\footnotesize
\centering
    \begin{tabular}{l r r r r r}
         \textbf{Graph} & \textbf{Nodes} & \textbf{Edges} &  \textbf{Communities} & \textbf{Largest com/ty} & \textbf{Protected} \\
         \hline
         Citeseer & 3,327 & 4,732 & 2 & 668 & 701\\
         Highschool & 156 & 1,437 & 3 & 32 & 85\\
         Pokec & 49,683 & 100,000 & 2 & 411 & 26,395\\
         Polblogs & 105 & 441 & - & - & 43\\
         Polbooks & 1,224 & 19,090 & - & -  & 636\\
    \end{tabular}
    \caption{Characteristics of graphs on which we experiment.}
    \label{tab:data sets}
\end{table}

All experiments for the three graphs with communities run once for each analysed community and each setting, whereas for the final two graphs they run once for each setting. Experiments are seeded and enough in number to make aggregate reports robust; in total, we investigate $(3+2+2\text{ communities})\cdot 6\text{ variations}+2\text{ graphs}\cdot 3\text{ variations}=48$ tasks.

\subsection{Compared approaches}\label{compared approaches}
In this subsection we outline promising existing and new approaches that improve the fairness of base graph filters, such as the ones outlined in the next subsection. All implementations are available as open source.\footnote{\url{https://github.com/maniospas/pygrank-f}} A smaller scale ablation study of our proposed NSGFF system's architectural choices is presented in Appendix~\ref{ablation study}.

\begin{itemize}
\item[] \textit{None.} Running the base graph filter without injecting any type of fairness awareness. This approach is both the baseline against which to compute the preservation of original posteriors, i.e., it corresponds to the ideal AUC assessment during community detection and to zero utility loss during diffusion.
\item[] \textit{LFPRO \citep{tsioutsiouliklis2020fairness}.} Near-optimal redistribution of ranks causing disparate impact. Unlike this work, which aims to influence posteriors through prior editing, LFPRO directly edits posteriors by redistributing excess scores between protected and non-protected group nodes to improve the prule while keeping the scores non-negative. The original posteriors are mostly preserved by making small incremental changes. To prevent numerical underflows that might prevent this approach from converging in graphs with many nodes, we repeat the gradual redistribution of scores up to a numerical tolerance of $10^{-12}$.
\item[] \textit{LFPRP \citep{tsioutsiouliklis2021fairness}.} This approach can only be applied on the personalized pagerank filter with column-wise adjacency matrix normalization. It evolves previous fair random walk heuristics \citep{rahman2019fairwalk} to redistributing excess inter-group node score transference during graph shifts from the protected group to the non-protected group and conversely. The redistribution within each group is weighted proportionally to the original posteriors. The weighing could also be uniform (known as the LFPRU approach \citep{tsioutsiouliklis2021fairness}), but this yields worse results across all experiments and we do not report it to make comparisons easier to parse. LFPRP achieves exact $\phi$-fairness (Subsection~\ref{fairness background}) for the adjusted node scores $r-(1-a)p$, where $r$ are its posteriors and $p$ the priors. In our experiments we assess the adjusted scores and set the value of $\phi$ as the equivalent of the target prule $=1$ described in Subsection~\ref{fairness background}.
\item[] \textit{Mult (baseline).} Applying the post-processing of Subsection~\ref{workaround} on base filters to remove the disparate impact of their posteriors. This serves both as a baseline to compare with other approaches and an ablation study that demonstrates gains of prior editing using the NSGFF approach presented below. 
\item[] \textit{FP \citep{krasanakis2020applying}.} Employing heuristic prior editing instead of this work's neural mechanism. The success of this approach hinges on its underlying assumptions being able to model optimization trajectories. We optimize the same utility loss as this work while applying the same filter transfer tricks (otherwise, this approach gets stuck in very shallow minima in some graphs). Best parameters are estimated using \textit{pygrank}'s black-box tuning \citep{krasanakis2022autogf} and set it to divide the parameter search range by two on each iteration and default other hyperparameters.
\item[] \textit{NSGFF (this work).} The neural surrogate graph filter fairness introduced in this work, optimized for the utility loss with ad-hoc regularization parameter $l_{reg}=\tfrac{\|q_0\|_1}{|\mathcal{V}|}$ (this is sufficient for Theorem~\ref{adjust theorem} as long as $\|q_0\|_1\geq 2\|q-q_0\|$ everywhere on some trajectory from $q_0$ to $q$). Following common graph neural network training practices, training employs the Adam optimizer with learning rate $0.01$ and default other hyperparameters \citep{kingma2014adam}, and the number of training epochs is determined by repeating parameter updates until the loss does not decrease for a patience of $100$ consecutive epochs. The number of neural layers $L$ and the filter transfer trick hyperparameter $\delta$ are tuned each time training takes place. To limit running time, we make this exploration coarser by searching among all combinations of $L\in\{3,\dots,9\}$ and $\delta_0\in\{0.1,1,10\}$ with shallow training that has patience of $5$ epochs, a proportional sped up learning rate of $0.1$, and an upper limit of $50$ epochs.
\end{itemize}

\subsection{Evaluation methodology}\label{methodology}
We now detail the methodology of assessing competing approaches. This consists of going through all $48$ filtering tasks outlined Subsection~\ref{graphs} and combining these with the $8$ base graph filters presented below and the $6$ compared approaches of Subsection~\ref{compared approaches} to mitigate the disparate impact of base filters. Since LFPRP can only be applied on personalized pagerank with asymmetric adjacency matrix normalization, experiments create a total of $48\text{ tasks}\cdot (6\text{ filters}\cdot 5\text{ approaches}+2\text{ filters}\cdot 6\text{ approaches})=2,016$ posterior signals. Each of these is assessed in terms of prule and, depending on the task, AUC or utility loss. To simplify presentation of results, we report an average of measure assessments for compared approaches per base filter and disparate impact mitigation approach. The average is weighted so that each graph is equally represented. We remind that the goal is not to find the best base filters, but to find the best disparate impact mitigation approach.

Experiments are conducted on the two popular families of graph filters outlined in Subsection~\ref{gsp}: personalized pagerank and heat kernels. We denote these as PPR\{a\}\{norm\} and HK\{t\}\{norm\} and create variations with propagation parameters $a=\{0.85, 0.9\}$ and $t=\{1,3\}$ and either symmetric or column-wise adjacency matrix normalization $norm\in\{\text{Sym, Col}\}$. 
Depending on the choice of filter type, propagation parameter, and adjacency matrix normalization, we study $2 \cdot 2 \cdot 2=8$ base filters. All filters are implemented to account up to their $20$-th polynomial term (this is their maximal receptive breadth).

\subsection{Experiment results}\label{results}
Community member recommendation and graph diffusion experiments are summarized in Figures~\ref{tab:community} and~\ref{tab:diff} respectively. First, we assert that original posteriors, i.e., the ones procured by filtering with no fairness awareness, are imbalanced between the protected group and the rest. This is indeed the case for community member recommendation, as indicated by the inadmissible (less than 80\%) prule of base graph filters at the second column. Disparate impact is not as prominent in diffusion experiments, where the unbiased randomization of priors lets base filters exhibit high prule by themselves. But prule is still not maximal, which indicates that there exist some, though not pervasive, graph structure biases that prevent perfect transfer of the priors' statistical parity to posteriors. 

These observations verify the existence of prior and structural biases, and that the former are more invasive. Hence, fairness-inducing interventions are needed to combat disparate impact. Looking at results, our proposed NSGFF approach achieves maximal prule while also better preserving the community member recommendation AUC than other approaches and inducing small utility loss. Furthermore, compared to LFPRP, which is the second best-performing approach in terms of AUC, our system is applied on a variety of filters. It also presents significant improvements compared to the heuristic prior editing of FP---which serves as motivation for this work.

\begin{table}[b]
\setlength{\tabcolsep}{2pt}
    \centering
    \begin{tabular}{l c c c c c c c c c c c c}
    ~ & \multicolumn{2}{c}{{None}} & \multicolumn{2}{c}{{LFPRO}} & \multicolumn{2}{c}{{LFPRP}} & \multicolumn{2}{c}{{Mult}} & \multicolumn{2}{c}{{FP}} & \multicolumn{2}{c}{{NSGFF}} \\
    \cmidrule(lr){2-3} \cmidrule(lr){4-5} \cmidrule(lr){6-7} \cmidrule(lr){8-9} \cmidrule(lr){10-11} \cmidrule(lr){12-13}
 ~ & AUC & prule & AUC & prule & AUC & prule & AUC & prule & AUC & prule & AUC & prule\\
 PPR0.85Col & 0.759  & 0.603  & 0.703  & 0.842  & 0.744 &  1.000     & 0.742 & 1.000  &  0.742  & 1.000   &  \textbf{0.752} &  1.000\\
 PPR0.90Col & 0.756  & 0.625  &  0.690  & 0.864 & 0.742  &  1.000  &   0.740  &  1.000  &  0.739  &  1.000  &  \textbf{0.751} & 1.000 \\
 HK1Col & 0.765  &  0.479   &  0.652  &  0.776  &  - & - &    0.750  &  1.000   &  0.750  &  1.000   &  \textbf{0.766} & 1.000\\
 HK3Col & 0.763  &   0.577  &  0.720  & 0.834  &  -  & -    & 0.749  &  1.000  &   0.749  &  1.000   &  \textbf{0.762}  &   1.000   \\
 PPR0.85Sym & 0.774  &  0.588  & 0.722 & 0.856  &  - & - & 0.750  &  1.000 &    0.755  &  1.000  &  \textbf{0.766} &  1.000 \\
 PPR0.90Sym & 0.773 & 0.619 & 0.730 & 0.884 & - & - & 0.749 & 1.000 & 0.750    & 1.000 & \textbf{0.763}  &  1.000\\
 HK1Sym & 0.774 & 0.445  &  0.664 & 0.763  &  - & - & 0.753  &   1.000   & 0.761   &  1.000  &  \textbf{0.766}  &  1.000 \\
 HK3Sym & 0.774 &  0.555  &  0.737 & 0.834  & - & - & 0.755  &   1.000  & 0.758  &  1.000  &  \textbf{0.766}  &  1.000 
    \end{tabular}
    \caption{Community member recommendation. Best fairness-aware AUC is bolded.}
    \label{tab:community}
\end{table}

\begin{table}[htpb]
\setlength{\tabcolsep}{2pt}
    \centering
    \begin{tabular}{l c c c c c c c c c c c c}
    ~ & \multicolumn{2}{c}{{None}} & \multicolumn{2}{c}{{LFPRO}} & \multicolumn{2}{c}{{LFPRP}} & \multicolumn{2}{c}{{Mult}} & \multicolumn{2}{c}{{FP}} & \multicolumn{2}{c}{{NSGFF}} \\
    \cmidrule(lr){2-3} \cmidrule(lr){4-5} \cmidrule(lr){6-7} \cmidrule(lr){8-9} \cmidrule(lr){10-11} \cmidrule(lr){12-13}
 ~ & $\mathcal{L}_{util}$ & prule & $\mathcal{L}_{util}$ & prule & $\mathcal{L}_{util}$ & prule & $\mathcal{L}_{util}$ & prule & $\mathcal{L}_{util}$ & prule & $\mathcal{L}_{util}$ & prule\\
 PPR0.85Col & 0.000   &  0.916  &   0.111   &  0.999   &  0.174 &    1.000   &  \textbf{0.041}  &   1.000  &   0.283  &   1.000  &   0.084  &   1.000\\
 PPR0.90Col & 0.000  &  0.923  &   0.094  &   0.999  &  0.146  &   1.000   &  \textbf{0.038}  &   1.000   &  0.215  &   1.000   &  0.073   &  1.000\\
 HK1Col & 0.000   &  0.907   &  152.596  & 0.993   &  - &   -   &  \textbf{0.047}   &  1.000   &  \textbf{0.047}  &   1.000  &   0.173   &  1.000\\
 HK3Col & 0.000  &   0.897   &  0.258   &  0.998  &   -  &  -  &   \textbf{0.053}   &  1.000   &  0.378  &  1.000  &   0.100   &  1.000  \\
 PPR0.85Sym & 0.000  & 0.934  &  0.051  &  0.999  &  - & - & 0.032  &  1.000  &  0.032  &  1.000  &  \textbf{0.024}  &  1.000\\
 PPR0.90Sym & 0.000  &  0.934  &  0.051  &  0.999  & - & - & 0.032   &  1.000  &   0.032   &  1.000  &  \textbf{0.026}  &  1.000\\
 HK1Sym & 0.000  &  0.934   &  73.384  &  0.997  & - & - & 0.032  &  1.000   &  0.032  &  1.000  &  \textbf{0.019}  &  1.000 \\
 HK3Sym & 0.000  &  0.926  &  0.128  &  0.999  &  - & - & 0.037  &  1.000   &  0.037  &  1.000  &  \textbf{0.027}  &  1.000 \\
    \end{tabular}
    \caption{Node value diffusion. Best fairness-aware utility loss is bolded.}
    \label{tab:diff}
\end{table}

Notably, the Mult baseline is a viable competitor against previous state-of-the-art. In fact, for diffusion tasks, where small reweighing suffices to fully mitigate disparate impact, it yields the smallest utility loss for asymmetric filtering compared to all other approaches. This baseline runs the risk of being less successful in debiasing diffusion with high disparate impact (Appendix~\ref{ablation study}), but its usage should by hereby considered in practical applications (Subsection~\ref{threats to validity}) and baselines of future work, especially since it preserves a similar portion of AUC compared to LFPRP. Nonetheless, NSGFF remains by far the best approach for all symmetric filtering and some asymmetric filtering tasks.

\section{Discussion}
In this section we list theoretical and practical properties to keep in mind when deploying our approach. We follow four threads. First, in Subsection~\ref{threats to applicability} we describe theoretical limits---mostly for graphs with too few nodes---on finding deep local optima for arbitrary objectives. Second, in Subsection~\ref{sec:robustness} we delve into the role of NSGFF's filter transfer trick in controlling posterior preservation and the robustness of local optima. Third, in Subsection~\ref{other objectives} we point to objectives other than posterior preservation. Fourth, in Subsection~\ref{threats to validity} we analyse the scalability of our approach and how this affects practical usage.

\subsection{Applicability criteria}\label{threats to applicability}
Going back to the generic prior editing framework of Section~\ref{optimizing posterior scores}, when the desired posterior loss is not non-quadratic (i.e., when the Hessian at one point can not form a good approximation of derivatives everywhere and consequently forms a large error $\epsilon_{\mathbb{H}}$), the radius of the domain from which ideal posteriors can attract original ones, as described in Theorem~\ref{adjust theorem} for adequate regularization, becomes dominated by the term: $$radius=0.5|\mathcal{V}|\tfrac{\lambda_{1} \min_v p[v]}{\lambda_{\max} \max_v p[v]}$$
We call this quantity \textit{optimization horizon radius}, as it indicates a minimal L2 distance between ideal (maybe locally if not globally optimal) posteriors and the ones produced by filters that, if not exceeded, always lets our methodology discover the former from the latter. That is, any posteriors within hyperspeheres of such radii around local optima will be attracted towards the ideal centers. If such hyperspheres are overlapping for multiple optima, it is uncertain which of those our approach will select. Whereas for distances greater than the the one indicated by the radii, it is unclear whether our approach will approximate optimal posteriors at all. 

An implicit assumption that drives the applicability of our posterior optimization framework ---and by extension the derived NSGFF system designed to mitigate disparate impact---is that original posteriors are close to the desired ones; {large enough optimization horizon radii allow greater edits to be applied on posteriors} while searching for optimal points without picking up trajectories that miss the latter. Sometimes, appropriate and potentially deeper edits can also be found for smaller radii (Subsection~\ref{sec:robustness}). 
Overall, radius values depend on the spectral characteristics of filters, the post-processing involved, and the number of graph nodes. When these create concerns over the applicability of our approach, we suggest validating the effectiveness of our framework instead of blindly applying it, for instance by following settings similar to our experimental methodology to assert that unsupervised losses are smaller than those of alternatives. Our analysis primarily targets connected graphs; for non-connected ones, the number of nodes when computing the radius should be replaced by the smallest among connected components.

In addition to understanding limitations for graphs with too few nodes, the above intuition can also explain why directly optimizing posteriors, for instance with LFPRO in experiments or NN in the ablation study of Appendix~\ref{ablation study}, is fundamentally flawed, as first theorized in the beginning of Section~\ref{optimizing posterior scores}. This mechanism can be understood as employing the identity filter $F(\hat{A})=I$ to transform some priors that are also posteriors a different (the base) graph filter. When all elements of the post-processing vector $p$ are equal (e.g., when L1 normalization is applied to posteriors), such approaches exhibit exceptionally large optimization horizon radii $0.5|\mathcal{V}|$, which are likely to encompass locally optimal posterior adjustments. But, at the same time, simple optimization does not respect any propagation mechanism, which means that many of the competing local optima candidates will not preserve the relations between adjusted posteriors that would have been imposed by the graph structure, i.e., many of those minima are shallow. 

\subsection{Using the filter transfer trick to control robustness}\label{sec:robustness}
Armed with the concept of the optimization horizon radius, we finalize the analysis the filter transfer trick introduced in Subsection~\ref{extending} and examine the impact of different values of the hyperparameter $\delta=\delta_0\max_{v}r_0[v]$ on the radii attracting posteriors to (asymmetric filtered or numerically robust) locally optimal points $r_{ns}$. To this end, we use Theorem's~\ref{trick opt} equivalent problem to find an optimization horizon radius and transcribe that back to the space of outputted surrogate posteriors:
\begin{align*}
&radius= 0.5|\mathcal{V}|\tfrac{\lambda_{1} \min_v \tfrac{p[v]}{\delta+r_0[v]}}{\lambda_{\max} \max_v \tfrac{p[v]}{\delta+r_0[v]}}\tfrac{1}{\delta+\max_v r_0[v]}
\geq 0.5\tfrac{\delta_0}{(\delta_0+1)^2\max_v r_0[v]}|\mathcal{V}|\tfrac{\lambda_{1} \min_v {p[v]}}{\lambda_{\max} \max_v {p[v]}}\quad
\\&\quad\geq  \tfrac{0.5\delta_0}{(\delta_0+1)^2}|\mathcal{V}|^2\tfrac{\lambda_1}{\lambda_{\max}}  \text{ for L1 normalization of }r_0
\end{align*}
Given that inequalities are inclusive (they can be equalities given appropriate conditions) we consider the last expression to be the optimization horizon radius when the filter transfer trick is employed. The radius grows quadratically with the number of graph nodes, which means that local minima are likely to be found, even for graphs with few nodes, though this can be deceptive in that the the same radius without the square is achieved when normalization snaps the maximum posterior to one. The radius for the underlying symmetric filtering task is always $\tfrac{\delta_0}{\delta_0+1}$ that of normal symmetric filtering (this holds true for any normalization that multiplies score vectors with scalars). 

A corollary of this analysis is that the filter transfer trick's hyperparameter $\delta_0$ controls how close local optima can be to ideal posteriors; small values encompass only local optima very close to posteriors, though they may also fail to find any. Given that our NSGFF approach hard-codes fairness constraints and is only interested in minimizing the utility loss, sufficiently small $\delta_0$ values that encompass at least one local minima effectively serve as mechanisms that reject larger yet locally optimal utility losses. Effectively, they shrink the hyperspheres around shallow local minima and thus prevent them from encompassing the original posteriors. 
For example, for symmetric filtering, the explored values $\delta_0\in\{0.1,1,10\}$ shrink optimization horizon radii to $0.091,0.5,0.909$ of those that would occur without the filter transfer trick, i.e., they range from only favoring posteriors very close to the original ones to maintaining almost the same radii.

Similarly, we explore the effect of Subsection~\ref{workaround}'s methodology, which hard-codes fairness constraints into our system. Given that all optimization horizon radius computations are multiplied with the term $\tfrac{\min_v p[v]}{\max_v p[v]}$, where $p$ is the post-processing vector, applying the disparate impact mitigation constraint effectively multiplies radii with the term:
$$\inf_{r\in\mathcal{R}}\min\big\{\tfrac{1-\phi_*(r)}{\phi_*(r)}\tfrac{\phi}{1-\phi}, \tfrac{\phi_*(r)}{1-\phi_*(r)}\tfrac{1-\phi}{\phi}\big\}=\inf_{r\in\mathcal{R}}\frac{\min\{prule_*(r),prule\}}{\max\{prule_*(r),prule\}}$$
where in this case $\mathcal{R}$ denotes the optimization domain or trajectory, $\phi$ and $prule$ are the ideal values of respective disparate impact quantifications we impose on our system, and $\phi_*(t)$ and $prule_*(t)$ compute these values for the filter's posteriors within the forward pipeline, i.e., before applying the constraint. The last two quantities can not be analytically derived but, when learning posteriors similar to the original ones as we do, we expect them to be similar to the corresponding computations of the original filter $\phi_*(r_0)$, $prule_*(r_0)$. Thus, when imposing the constraint $prule=1$, as we did to fully mitigate disparate impact, the optimization horizon radius is approximately multiplied with $prule_*(r_0)$ computed for base filters (this corresponds to the None approach in experiments). This analysis reveals that it may be challenging to debias graph filtering with disparate impact very close to zero.


\subsection{Objectives other than posterior preservation}\label{other objectives}
In experiments, our surrogate graph neural network model succeeds in tightly approximating original posteriors (as indicated by the small utility loss) while mitigating disparate impact from filter score fairness, and should therefore be preferred for this objective. Small utility losses in general correspond to high community recommendation quality too. However, following posteriors less tightly could produce other business benefits.

Characteristically, fairness sometimes improves AUC in community recommendation experiments, as seen for its higher average value for NSGFF compared to vanilla HK1Col filtering. In such cases, by merit on being more inclusive, fairness-aware graph filtering identifies more representative communities. Indeed, like in the real world, the random prior selection methodology used in experiments has a chance of excluding highly relevant areas of the graph by not exhibiting non-zero priors near them. Consequently, inclusivity during filtering lets it account for relevant faraway areas in the recommendations. This argument has previously been voiced by \cite{stoica2020seeding}, although in our case we infer the more diverse prior node values instead of sampling them from the real world.

Thus, there could be interest in approaches that follow posteriors less tightly in favor of other objectives. 
Future works can accommodate such criteria, for instance via a supervised instead of utility losses, or training schemes that progressively move away from the original posteriors while maintaining local optimality. Such research could extend our local universal approximation results to families of graph neural networks for node classification, as described in Appendix~\ref{towards universal}. In addition to other objective/loss components, our theoretical framework is made universally applicably to twice differentiable objectives only by enriching it with a sufficiently large L1 regularization term. In experiments, we obtained near-identical results when omitting this term, which is to be expected given how small its penalties become relative to other loss components for graphs with hundreds or more nodes.

\subsection{Scalability limitations}\label{threats to validity}
Before closing this work, we point out that the computation cost of running our approach can be prohibitive for large graphs. Training the graph neural network architecture typically requires up to a few thousand filter runs, and the hyperparameter search for its ideal depth repeats several training procedures. These costs are no greater than what would be expected of other graph neural networks in the literature addressing different tasks. Furthermore, they scale linearly with the cost of running one graph filter and can be sped up with distributed or parallel computing. At the same time, the added societal value of making graph filtering fairness-aware can be worth the filtering delay or infrastructure cost.

On the other hand, compared to the small cost of running simple filters, our approach is not well-suited to processing large graphs with millions of nodes, as it requires thousands of filter reruns. For example, the running time of our system's implementation (including training on-the-fly and hyperparameter exploration) for the Citeseer graph on a 2.6GHz system with DDR3 RAM requires several minutes, but a single filter runs in less than 20 milliseconds. Given that we even go beyond this scope and conduct large-scale benchmarking, our investigation is limited to graphs with fewer (thousands instead of mullions of) nodes and lightweight hyperparameter search. Still, given enough resources, we expect similar or better results to be obtained, even for graphs with more nodes, whose optimization horizon radii are larger. 

GPU framework implementations of sparse matrix multiplications needed by graph shifts that do not exhibit parallelization gains are partly to blame for the lack of speed. We expect future breakthroughs in related technologies to help better scale our approach. For the time being, and when processing graphs with millions of edges, there is still merit to devising heuristics, like LFPRP or even the Mult baseline, that may find solutions shallower but comparable to ours. Given the increasing fairness concerns of applying artificial intelligence in many aspects of the real world, even naive alternatives to our approach are preferable to taking no measures against disparate impact when that is a concern.


\section{Conclusions and Future Work}
In this work we explored the concept of editing graph signal priors to locally optimize fairness-aware graph filter posterior objectives. To this end, we introduced a framework for filter-aware universal approximation of local posterior objective minima. In this framework, we set fairness-aware objectives, and implemented the resulting system via a graph neural network architecture to be used as a surrogate of ideal graph filtering. The architecture conducts new optimization and hyperparameter search at runtime for new priors and filters. We also designed output activations so that, when posteriors are non-negative, ensure usability even under non-symmetric adjacency matrix normalization and numerically unstable objectives. 
We finally experimented on real-world graphs with various graph filters and signal priors, and used our approach to mitigate disparate impact while producing higher quality posterior scores than competing approaches, in terms of either AUC or an unbiased utility loss.
\par
Promising applications of this work could involve designing more fair or powerful decoupled graph neural network architectures, where feature extraction and propagation with graph filters are performed separately. Future research can also improve our approach via more informed (e.g., learned) filter transfer tricks for non-symmetric filtering, as well as algorithmic speed-ups to training, hyperparameter exploration, and training. Finally, we recognize the scientific validity of devising evaluation objectives for graph filtering that are not affected by the biases being mitigated, as we did for our selected utility loss, and using them to assess existing approaches.

\acks{This work received funding by the European Union under contract number GA-101070285 MAMMOth.}


\bibliography{main}{}

\clearpage
\appendix

\section{Towards Universal Approximation on the Node Domain}\label{towards universal}
The theoretical analysis of this work reaches beyond fairness-aware graph filtering and presents the first steps towards universal approximation theorems in graph neural networks for node attribute prediction (e.g., classification). Recent research grapples with the expressive power of node representations \citep{xu2018powerful,chen2019equivalence,becigneul2020optimal,azizian2020expressive}, but this concept is weaker than the universal approximation exhibited by traditional neural networks \citep{li2022expressive}. In this work, we express universal approximation of node scores \textit{in one specific graph each time} and guarantee (local) optimization under the existence of both sufficiently good filters to be used and a full encoding of graph signals involved in objective calculations in the priors. 

Thanks to the above restrictions, our analysis can universally approximate graph signal posterior objectives in the traditional sense. Furthermore, it indicates that practices injecting additional (e.g., random) attributes to graph neural network features do not necessarily need to capture node identities, as \cite{li2022expressive} hypothesize, but only classes of nodes with different desired posteriors. For example, two nodes with the same features or representations would be allowed if they also had the same prediction targets, and despite whether this phenomenon would be a manifestation of limited expressive power.


At the same time, we provide a first---to our knowledge---formal understanding of \textit{why} graph neural networks work so well compared to optimization that does not acknowledge the graph structure. In theory, for tasks where data are abundant, graph shift operators could be approximated via neural network components (e.g., that mimic matrix decomposition) so that posterior objectives could be optimized via traditional deep architectures, but this is rarely sufficient in practice \citep{hoang2021revisiting}. 
The discussion of Subsection~\ref{threats to applicability} presents a more delicate view than brute force optimization; the graph structure plays the role of a selection mechanism that constraints the optimization to local minima of only certain types. Thus, if there exist correlations between structural characteristics captured by graph filters and deep minima, shallow minima will be avoided by graph neural networks.

Kindred principles have led to the adoption of personalized pagerank under symmetric adjacency matrix normalization in lieu of regularization mechanisms that impose similar predictions between neighbors \citep{zhou2003learning,huang2020combining}. Our work explains why this is needed in terms of optimization theory; given that the imposed regularization corresponds to qualitative predictions by itself, neural preprocessing can appropriately adjust its predictions. Furthermore, we generalize to any kind of graph filter at the role of the regularizer, given that it works as an adequately good predictor by itself. The need for expressing qualitative predictions despite not encountering these during training is also reflected in graph neural network architectures like GCNII \citep{chen2020simple}, which artificially impose low-pass filtering by reducing the importance of many-hop graph diffusion.

Finally, a similar theory to ours is presented by \cite{hoang2021revisiting}, who point out that existing graph neural network research focuses on low-pass filters and datasets with small eigenvalues and that new approaches should consider other variations too. In this work, we support generalized filtering, but also discover pitfalls corresponding to small optimization horizon radii: some types of high-pass filtering (e.g., combining high-pass filters and graphs with small eigenvalues) admit too small radii for a small number of nodes and are therefore only applicable to graphs with many nodes.



\section{Ablation Study}\label{ablation study}
Subsection~\ref{results} demonstrates the ability of our proposed graph neural network framework to act as a generic surrogate of fairness-aware graph filtering. However, the NSGFF system is built by combining several components. In this appendix, we conduct an empirical ablation study to showcase that each of these components proactively circumvents issues that arise from misleadingly similar architectures. 

For demonstration purposes, we conduct a small scale comparison between our approach and variations that give an idea of how things can go wrong when optimizing the posterior objectives of symmetric filtering. The comparison is made on a diffusion task defined on the {Citeseer} graph by assigning either values $1$ or uniformly random values from the range $[0,1]$ to the first community's members and diffusing those. We investigate the following variations of NSGFF:

\begin{itemize}
    \item[] \textit{Mult (baseline).} The posterior rebalancing approach described in Subsection~\ref{compared approaches}. This effectively removes the learning aspect of our approach.
    \item[] \textit{NN (baseline).} Directly editing posteriors by removing the graph propagation step of NSGFF to turn it into a multilayer perceptron. This strategy retains the same output activation functions, including those imposing fairness constraints, is trained on the same objective, and runs the same neural depth hyperparameter exploration. 
    \item[] \textit{APPNP \citep{klicpera2018predict}.} Adopting a fixed personalized pagerank filter with parameter $a=0.9$ and $10$ (instead of $20$ that we otherwise use in this work) propagation steps as NSGFF's filter. Selecting a filter different than the one our framework aims to substitute is of unknown efficacy, especially since even the receptive breadth is different. Future works may be interested in investigating different filters of similar structural characteristics but greater optimization horizon radii, but this prospect lies outside the scope of our research.
    \item[] \textit{NSGFF (this work).} Our proposed system, as summarized in Subsection~\ref{compared approaches}.
\end{itemize}

Results from making the particular diffusion task fairness-aware are presented in Table~\ref{tab:ablation}. The full architecture does not always arrive at smaller utility losses compared to cruder variations, but this is not unexpected, given that the parameter $\delta_0$ controlling the optimization horizon radius is selected from only a small range of options and may let original posteriors lie within the radii of multiple local minima, including shallower ones. Still, all utility losses of our approach reside in deep minima, i.e., they are near-best compared to alternatives. Conversely, variations of our scheme run the risk of ending up with much higher losses. 
\clearpage
\begin{table}[!htb]
\setlength{\tabcolsep}{2pt}
    \centering
    \begin{tabular}{l c c c c c c c c c c c c}
    ~ & \multicolumn{2}{c}{{None}} & \multicolumn{2}{c}{{Mult}} & \multicolumn{2}{c}{{NN}} & \multicolumn{2}{c}{{APPNP}} & \multicolumn{2}{c}{{NSGFF}} \\
    \cmidrule(lr){2-3} \cmidrule(lr){4-5} \cmidrule(lr){6-7} \cmidrule(lr){8-9} \cmidrule(lr){10-11} 
 ~ & $\mathcal{L}_{util}$ & prule & $\mathcal{L}_{util}$ & prule & $\mathcal{L}_{util}$ & prule & $\mathcal{L}_{util}$ & prule & $\mathcal{L}_{util}$ & prule \\
\multicolumn{11}{c}{Diffusion of random values}\\
PPR0.85Sym & 0.000   & 0.138  & 0.689   & 1.000   & 0.623  &   1.000  &   0.467  &   1.000   &  \textbf{0.356}   &  1.000\\
PPR0.90Sym & 0.000   &  0.153  &   0.619  &   1.000   &  0.555   &  1.000   &  \textbf{0.336} &    1.000  &   0.343   &  1.000\\
HK1Sym & 0.000    & 0.051   &  1.860  &   1.000 &    1.788  &   1.000  &   0.598  &   1.000 &   \textbf{0.488}  &   1.000\\
HK3Sym & 0.000  &   0.116  &   0.818  &   1.000   &  0.751  &   1.000   &  0.376   &  1.000  &   \textbf{0.365}   &  1.000 \\
\multicolumn{11}{c}{Diffusion of binary values}\\
PPR0.85Sym & 0.000  &   0.134  &   0.720  &   1.000   &  0.656   &  1.000 &    0.399  &   1.000  &   \textbf{0.369}  &   1.000\\
PPR0.90Sym & 0.000  &   0.138   &  0.698   &  1.000   &  0.636  &   1.000   &  \textbf{0.361}  &   1.000   &  0.366  &   1.000\\
HK1Sym & 0.000  &  0.039   &  2.477   &  1.000   &  2.406  &   1.000 &    0.732   &  1.000   &  \textbf{0.648}  &   1.000 \\
HK3Sym & 0.000   &  0.107   &  0.901   &  1.000   &  0.836  &   1.000  &   0.398   &  1.000  &   \textbf{0.383}  &   1.000
 
    \end{tabular}
    \caption{Diffusion in the Citeseer graph. Best fairness-aware utility loss is bolded.}
    \label{tab:ablation}
\end{table}

\clearpage
\section{Proofs}\label{proofs}
In this appendix we present proofs of the theoretical results supporting our work.\\

\noindent\textbf{Lemma~\ref{robustness}}
\textit{Let $F(\hat{A})$ be a positive definite graph filter and $p$ a post-processing vector. If $f(r)$ is a $\tfrac{\lambda_1 \min_v p[v]}{\lambda_{\max} \max_v p[v]}$-optimizer of a differentiable loss $\mathcal{L}(r)$ over graph signal domain $\mathcal{R}\subseteq\mathbb{R}^{|\mathcal{V}|}$, where $\lambda_1,\lambda_{\max}>0$ are the smallest positive and largest eigenvalues of $F(\hat{A})$ respectively, updating graph signals per the rule:
\begin{equation}\label{update}
\begin{split}
    &\frac{\partial q(t)}{\partial t}=f(r(t))
    \\&r(t)=\text{diag}(p)F(\hat{A})q(t)
\end{split}
\end{equation}
asymptotically leads to the loss to local optimality if posterior updates are closed in the domain, i.e. $r(t)\in \mathcal{R}\,\forall t\in[0,T]\Rightarrow \text{diag}(p)F(\hat{A})\int_0^{T}f\big(r(t)\big)dt\in\mathcal{R}$.}
~\\\begin{proof}
For non-negative posteriors $r=\text{diag}(p)F(\hat{A})q(t)$, and non-zero loss gradients, the Cauchy-Shwartz inequality in the bilinear space $\langle x,y\rangle=x^TF(\hat{A})y$ determined by the positive definite graph filter $F(\hat{A})$ yields:
\begin{align*}
    &  \frac{d \mathcal{L}\big(r(t)\big)}{dt}
    \\&\quad=\Big(\nabla \mathcal{L}\big(r(t)\big)\Big)^T \text{diag}(p) F(\hat{A}) \frac{d q(t)}{dt}
    \\&\quad=\Big(\nabla \mathcal{L}\big(r(t)\big)\Big)^T \text{diag}(p) F(\hat{A})f\big(r(t)\big)
    \\&\quad=\Big(\nabla \mathcal{L}\big(r(t)\big)\Big)^T \text{diag}(p) F(\hat{A})\big( -\nabla \mathcal{L}\big(r(t)\big)+(f\big(r(t)\big)+\nabla\mathcal{L}\big(r(t)\big))\big)
    \\&\quad\leq -\min_v p[v]\Big(\nabla \mathcal{L}\big(r(t)\big)\Big)^T F(\hat{A})\nabla \mathcal{L}\big(r(t)\big)
    +(\nabla \mathcal{L}(r))^T\text{diag}(p)F(\hat{A})\Big(f\big(r(t)\big)+\nabla\mathcal{L}\big(r(t)\big)\Big)
    \\&\quad\leq -\lambda_1\min_v p[v]\|\nabla \mathcal{L}\big(r(t)\big)\|^2
    +\lambda_{\max}\big\|f\big(r(t)\big)+\nabla\mathcal{L}\big(r(t)\big)\big\|\|\text{diag}(p)\nabla \mathcal{L}\big(r(t)\big)\|
    \\&\quad< -\min_v p[v]\lambda_1\big\|\nabla \mathcal{L}\big(r(t)\big)\big\|^2
    +\lambda_{\max}\tfrac{\lambda_1 \min_v p[v]}{\lambda_{\max} \max_v p[v]}\big\|\nabla \mathcal{L}\big(r(t)\big)\big\|\max_v p[v]\big\|\nabla \mathcal{L}\big(r(t)\big)\big\| 
    \\&\quad= 0
\end{align*}
Therefore, 
the loss asymptotically converges to a locally optimal point.
\end{proof}

\noindent\textbf{Lemma~\ref{anyedit}}
\textit{Let us consider graph signal priors $q_0$, positive definite graph filter $F(\hat{A})$ with largest and smallest eigenvalues $\lambda_{\max},\lambda_1$, post-processing vector $p$, differentiable loss function $\mathcal{L}(r)$ with at least one minimum in domain $\mathcal{R}\subseteq\mathbb{R}^{|\mathcal{V}|}$, and a differentiable graph signal generation function $\mathcal{M}:\mathbb{R}^K\to\mathbb{R}^{|\mathcal{M}|}$ for which there exist parameters $\theta_0$ satisfying $\mathcal{M}(\theta_0)=q_0$ and $\text{diag}(p)F(\hat{A})\mathcal{M}(\theta)\in\mathcal{R}$. If, for any parameters $\theta$, its Jacobian $\mathbb{J}_\mathcal{M}(\theta)$ has linearly independent rows and satisfies:
\begin{align*}
    &\big\|E(\theta)\nabla \mathcal{L}(r)\big\|< \tfrac{\lambda_1 \min_v p[v]}{\lambda_{\max} \max_v p[v]}\|\nabla \mathcal{L}(r)\|\quad\text{ for }\nabla \mathcal{L}(r)\neq \textbf{0}\\
    &E(\theta) = \mathcal{I}-\mathbb{J}_\mathcal{M}(\theta)(\mathbb{J}_\mathcal{M}^T(\theta)\mathbb{J}_\mathcal{M}(\theta))^{-1}\mathbb{J}_\mathcal{M}^T(\theta)\\
    &r(\theta) = \text{diag}(p)F(\hat{A})\mathcal{M}(\theta)
\end{align*}
then there exist parameters $\theta_{\infty}$ that make $\mathcal{L}\big(r(\theta_\infty)\big)$ locally optimal.}

~\\\begin{proof}
Let us consider a differentiable trajectory for parameters $\theta(t)$ for times $t\in[0,\infty)$ that starts from $\theta(0)=\theta_0$ and (asymptotically) arrives at $\theta_\infty=\lim_{t\to\infty} \theta(t)$. For this trajectory, it holds that $\tfrac{d \mathcal{M}(\theta(t))}{dt}=\mathbb{J}_\mathcal{M}\big(\theta(t)\big)\frac{d\theta(t)}{dt}$. Then, let us consider the posteriors $r(t)=\text{diag}(p)F(\hat{A})\mathcal{M}\big(\theta(t)\big)$ arising from the graph signal function $\mathcal{M}\big(\theta(t)\big)$ at times $t$, as well the least square problem of minimizing the projection of the loss's gradient to the row space of $\mathbb{J}_\mathcal{M}\big(\theta(t)\big)$: $$\text{ minimize }\big\|\nabla \mathcal{L}\big(r(t)\big)-\mathbb{J}_\mathcal{M}\big(\theta(t)\big)x(t)\big\|$$
The closed form solution to this problem can be found by: $$x(t)=\big({\mathbb{J}_\mathcal{M}}^T\big(\theta(t)\big)\mathbb{J}_\mathcal{M}\big(\theta(t)\big)\big)^{-1}\mathbb{J}_\mathcal{M}^T\big(\theta(t)\big)\nabla \mathcal{L}\big(r(t)\big)$$
Thus, as long as $q(t)=\mathcal{M}\big(\theta(t)\big)$ (Proposition I), the theorem's precondition for posteriors $r(t)=\text{diag}(p)F(\hat{A})q(t)$ and $\nabla \mathcal{L}\big(r(t)\big)\neq \textbf{0}$ can be written as:$$\big\|\nabla \mathcal{L}\big(r(t)\big)-\mathbb{J}_\mathcal{M}\big(\theta(t)\big)x(t)\big\|<  \tfrac{\lambda_1 \min_v p[v]}{\lambda_{\max} \max_v p[v]}\|\nabla \mathcal{L}\big(r(t)\big)\|$$
Hence, if we set $x(t)$ as parameter derivatives:
$$\tfrac{d \theta(t)}{dt}=x(t)\Rightarrow \Big\|\nabla \mathcal{L}\big(r(t)\big)-\tfrac{d \mathcal{M}(\theta(t))}{dt}\Big\|<  \tfrac{\lambda_1 \min_v p[v]}{\lambda_{\max} \max_v p[v]}\big\|\nabla \mathcal{L}\big(r(t)\big)\big\|$$
which means that, on the selected parameter trajectory, $\tfrac{\mathcal{M}(\theta(t))}{dt}$ is a $\tfrac{\lambda_1 \min_v p[v]}{\lambda_{\max} \max_v p[v]}$-optimizer of the loss function. As, such, from Lemma~\ref{robustness} it discovers locally optimal posteriors.
\par
As a final step, we now investigate the priors signal editing converges at. To do this, we can see that the update rule leads to the selection of priors $q(t)$ at times $t$ for which:
\begin{align*}
    &\tfrac{d q(t)}{dt}=\tfrac{d \mathcal{M}(\theta(t))}{dt}
    \\&\Rightarrow \lim_{t\to\infty} q(t)=q(0)+\int_{t=0}^{\infty} \tfrac{\mathcal{M}(\theta(t))}{dt} dt = \mathcal{M}\big(\theta(t_{\infty})\big)- \mathcal{M}\big(\theta(0)\big)=\mathcal{M}\big(\theta(t_{\infty})\big)
\end{align*}
\par
We can similarly show that (Proposition I) holds true. Hence, there exists an optimization path of prior editing parameters (not necessarily the same as the one followed by the optimization algorithm used in practice) that arrives at the edited priors $\mathcal{M}(\theta_\infty)$ for some parameters $\theta_\infty=\theta(t_{\infty})$ that let posteriors exhibit local optimality.
\end{proof}

\noindent\textbf{Theorem~\ref{mlp}}
\textit{Let us consider graph signal priors $q_0$, positive definite graph filter $F(\hat{A})$ with largest and smallest eigenvalues $\lambda_{\max},\lambda_1$ respectively, post-processing vector $p$, and 
twice differentiable loss function $\mathcal{L}(r)$, with some unique minimum at unknown posteriors $r=r_\infty$ of a domain $\mathcal{R}\subseteq\mathbb{R}^{|\mathcal{V}|}$, and whose Hessian matrix $\mathbb{H}_\mathcal{L}(r)$ linearly approximates its gradients $\nabla\mathcal{L}(r)$ within the domain with error at most $\epsilon_\mathbb{H}$. Let us also construct a node feature matrix $H^{(0)}$ whose columns are graph signals (its rows $H^{(0)}[v]$ correspond to nodes $v$). If $q_0$ and all graph signals involved in the calculation of $\mathcal{L}(r)$ are columns of $H^{(0)}$, and it holds that: $$\tfrac{\lambda_{1} \min_v p[v]}{\lambda_{\max} \max_v p[v]}\|\nabla\mathcal{L}(r)\|>2\epsilon_\mathbb{H}\|r-r_\infty\|$$
for any $r\in\mathcal{R}\setminus\{r_\infty\}$, then for any $\epsilon_\infty>0$ there exists a deep neural network architecture $$H^{(\ell)}=\phi_\ell(H^{(\ell-1)}W^{(\ell)}+b^{(\ell)})$$ with activation functions 
$$\phi_\ell(x)=\{x\text{ for }\ell=L,relu(x)\text{ otherwise}\}$$ learnable weights $W^{(\ell)}$, and biases $b^{(\ell)}$ for layers $\ell=1,2,\dots,L$ with depth $L\geq 4$, in which using the output of the last layer $H^{(L)}$ as priors yields posteriors arbitrarily close to the local minimum $\|diag(p)F(\hat{A})H^{(L)}-r_\infty\|<\epsilon_\infty$. Additionally, layers other than the last can have output columns equal to the number of $\text{columns of }H^{(0)}+2$.
}
~\\\begin{proof}
Let $\mathbb{H}_\mathcal{L}(r)$ be the Hessian matrix of $\mathcal{L}(r)$ with respect to node values of the graph signal $r\in\mathcal{R}$ of a domain $\mathcal{R}$. Given that its rank is $K(r)$, there exists a decomposition $\mathbb{H}_\mathcal{L}(r)=J(r)D(r)$ with fixed inner dimensions $K_{\max}=\max_{r\in\mathcal{R}} K(r)$. When $K(r)=K_{\max}$, the decomposition even becomes unique for the particular $r$.   

We now select a prior editing/generation model $\mathcal{M}(\theta)$ with linearly independent rows and a Jacobian $\mathbb{J}_\mathcal{\mathcal{M}}(\theta)$ that is a deep neural network approximating well the first of the decomposition elements. That is, for any chosen small enough (\textit{but non-negligible}) constant $\epsilon_\mathbb{J}>0$ within the domain $\mathcal{R}$ it holds that:
$$\big\|\mathbb{J}_\mathcal{\mathcal{M}}(\theta)-J(r)\big\|\leq\epsilon_\mathbb{J}$$ 
Throughout this proof, $r$ is a function of $\theta$, but we do not write $r(\theta)$ for simplicity. The Jacobian does not need to have linearly independent rows; small perturbations (with cumulative effect that, when added to the actual approximation error, makes it smaller than $\epsilon_\mathbb{J}$) can be injected in each row to make the rows of its approximation linearly independent.
Additionally, we use the Hessian as a linear estimator of gradients $\nabla \mathcal{L}(r_\infty)$ per:
$$\|\nabla\mathcal{L}(r_\infty)-\nabla\mathcal{L}(r)-\mathbb{H}_\mathcal{L}(r)(r_\infty-r)\|\leq\epsilon_\mathbb{H}\|r-r_\infty\|$$
We now rewrite the loss's gradient as a linear transformation of a factor $\mathcal{L}'(r)$ with an error term:
$$\nabla\mathcal{L}(r)=\mathbb{J}_\mathcal{\mathcal{M}}(\theta)\mathcal{L}'(r)+\epsilon(r)$$
where $\mathcal{L}'(r)=\mathbb{J}_\mathcal{\mathcal{M}}^T(\theta)\big(\mathbb{J}_\mathcal{\mathcal{M}}(\theta)\mathbb{J}_\mathcal{\mathcal{M}}^T(\theta)\big)^{-1}\nabla\mathcal{L}(r_\infty)+D(r)(r-r_\infty)$, and the error term admits a small bound:
$$\|\epsilon(r)\|\leq \epsilon_\mathbb{H}\|r-r_\infty\|+\epsilon_\mathbb{J}\|D(r)\|\|r-r_\infty\|\leq \sup_{r\in\mathcal{R}}\|r-r_\infty\|(\epsilon_\mathbb{H}+\sup_{r\in\mathcal{R}}\|D(r)\|\epsilon_\mathbb{J})$$
At this point, we choose $\epsilon_\mathbb{J}=\tfrac{\epsilon_\mathbb{H}}{\sup_{r\in\mathcal{R}}\|D(r)\|}$, which yields $\|\epsilon(r)\|\leq 2\epsilon_\mathbb{H}\|r-r_\infty\|$

Applying the linear estimation obtained above on the quantities of Lemma~\ref{anyedit} we obtain:
\begin{align*}
&\big\|E\nabla \mathcal{L}(r)\big\|=\|\nabla \mathcal{L}(r)-\mathbb{J}_\mathcal{M}(\theta)(\mathbb{J}_\mathcal{M}^T(\theta)\mathbb{J}_\mathcal{M}(\theta))^{-1}\mathbb{J}_\mathcal{M}^T(\theta)\big(\mathbb{J}_\mathcal{M}(\theta) \mathcal{L}'(r)+\epsilon(r)\big)\big\|
\\&=\|\nabla \mathcal{L}(r)-\mathbb{J}_\mathcal{M}(\theta)\mathcal{L}'(r)+\mathbb{J}_\mathcal{M}(\theta)(\mathbb{J}_\mathcal{M}^T(\theta)\mathbb{J}_\mathcal{M}(\theta))^{-1}\mathbb{J}_\mathcal{M}^T(\theta)\epsilon(r)\|
\\&=\|\epsilon(r)-\mathbb{J}_\mathcal{M}(\theta)(\mathbb{J}_\mathcal{M}^T(\theta)\mathbb{J}_\mathcal{M}(\theta))^{-1}\mathbb{J}_\mathcal{M}^T(\theta)\epsilon(r)\|
=\big\|E \epsilon(r)\big\|
\end{align*}
However, the matrix $\mathcal{I}-E=\mathbb{J}_\mathcal{M}(\theta)(\mathbb{J}_\mathcal{M}^T(\theta)\mathbb{J}_\mathcal{M}(\theta))^{-1}\mathbb{J}_\mathcal{M}^T$ is indempondent (that is, it holds $(\mathcal{I}-E)(\mathcal{I}-E)=\mathcal{I}-E$) and therefore its eigenvalues are either $0$ or $1$. Thus:
$$\|E\nabla \mathcal{L}(r)\big\|\leq\|\epsilon(r)\|\leq 2\epsilon_\mathbb{H}\|r-r_\infty\|<\tfrac{\lambda_{1} \min_v p[v]}{\lambda_{\max} \max_v p[v]}\|\nabla\mathcal{L}(r)\|$$
As a final step, and given that we showed the existence of a (deep neural network) function, $\mathbb{J}_\mathcal{\mathcal{M}}(\theta)$, we integrate it to retrieve $\mathcal{M}(\theta)$ and then approximate the integral with a new deep neural network. Given the universal approximation theorem's form described by \cite{kidger2020universal} the latter can be the architecture described by this theorem, used to produce error less than $\epsilon_\infty\tfrac{\lambda_1 \min_v p[v]}{\lambda_{\max}\max_v p[v]}$. In this case: \begin{align*}&\|diag(p)F(\hat{A})H^{(L)}-r_\infty\|=\|diag(p)F(\hat{A})H^{(L)}-diag(p)F(\hat{A})\mathcal{M}(\theta_\infty)\|
\\&\quad\leq\|diag(p)F(\hat{A})\|\|H^{(L)}-\mathcal{M}(\theta_\infty)\|<\|\epsilon_\infty\|
\end{align*}We could not immediately assume the existence of that final network, because we needed to verify that the Jacobian's properties for the function it approximates and ``hide'' methods of tracking equal partial derivatives of different nodes behind the term $\epsilon_\mathbb{J}$.
\end{proof}


\noindent\textbf{Theorem~\ref{adjust theorem}}
\textit{For any twice differentiable loss $\mathcal{L}(r)$, there exists sufficiently large enough parameter $l_{reg}\in\big[0,2\tfrac{\sup_q\|q-q_0\|}{|\mathcal{V}|}\big]$ such that the loss regularization of Equation~\ref{loss adjustment} satisfies the properties needed by Theorem~\ref{mlp} within the optimization domain $$\mathcal{R}=\{r_\infty\}\cup\bigg\{r:\|r-r_\infty\|<0.5\max\big\{|\mathcal{V}|,\tfrac{\|\nabla\mathcal{L}(r)\|}{\epsilon_\mathbb{H}}\big\}\tfrac{\lambda_{1} \min_v p[v]}{\lambda_{\max} \max_v p[v]}\bigg\}$$where $r_{\infty}$ is the ideal posteriors optimizing the regularized loss. If the second term of the $\max$ is selected, it suffices to have no regularization $l_{reg}=0$.}

~\\\begin{proof}
For ease of notation, during this proof we define $w=0.5\tfrac{\lambda_{1} \min_v p[v]}{\lambda_{\max} \max_v p[v]}$. 
Without loss of generality, consider $sgn$ to be a closure of the sign function on the optimization trajectory. If $\tilde{\epsilon}_\mathbb{H}$ and ${\epsilon}_\mathbb{H}$ are the maximum derivative approximation error of $\tilde{\mathcal{L}}(r)$ and $\mathcal{L}(r)$ respectively, it suffices to select $l_{reg}>0$ such that: 
\begin{align*}
&w\|\nabla\mathcal{L}(r)\|+wl_{reg}|\mathcal{V}| > (\epsilon_\mathbb{H}+l_{reg})\sup_{r}\|r-r_\infty\|
\\&\quad\Leftrightarrow l_{reg}(w|\mathcal{V}|-\sup_{r}\|r-r_\infty\|) > \epsilon_\mathbb{H}\sup_{r}\|r-r_\infty\|-w\|\nabla\mathcal{L}(r)\|
\\&\quad\Leftarrow |\mathcal{V}|w>\sup_{r}\|r-r_\infty\|\text{ or }\epsilon_\mathbb{H}\sup_{r}\|r-r_\infty\|<w\|\nabla\mathcal{L}(r)\|
\end{align*}
~\\where the first case needs to come alongside the condition $$l_{reg}>\tfrac{\sup_{r}\|r-r_\infty\|}{w|\mathcal{V}|-\sup_{r}\|r-r_\infty\|}\Leftarrow l_{reg}\Leftarrow l_{reg}=\tfrac{\sup_{r}\|r-r_\infty\|}{w|\mathcal{V}|}\geq \tfrac{2w\|q-q_0\|}{w|\mathcal{V}|}$$
but in the second case it suffices to select any $l_{reg}\geq 0$. Given an appropriate selection, we obtain the necessary criterion: $$w\|\tilde{\mathcal{L}}(r)\|\geq w\|\nabla\mathcal{L}(r)\|+wl_{reg}|\mathcal{V}|> (\epsilon_\mathbb{H}+l_{reg})\sup_{r}\|r-r_\infty\|\geq \tilde{\epsilon}_\mathcal{H}\|r-r_\infty\|$$
\end{proof}

\noindent\textbf{Theorem~\ref{trick opt}}
\textit{Let us consider the case where priors and filter parameters are all non-negative, and asymmetric adjacency matrix normalization yields positive edge weights. Equation~\ref{transform} can express locally optimal asymmetric filtering posteriors, which can be found by applying the graph neural architecture of this section on the equivalent symmetric filter to optimize the loss $\mathcal{L}(f_{\delta}(r))$ from node features $H^{(0)}$ extended with a fourth dimension $H^{(0)}[v,3]=r_{ns0}[v]$ and $6$ neural layer dimensions. The optimization tightness of the symmetric filter is the same as if we minimized $\mathcal{L}\big(diag\big(\tfrac{p}{\delta+r_0}\big)F(\hat{\mathcal{A}})q\big)$.}
~\\\begin{proof}
    We first define the receptive breadth of filters $F(\hat{A})$ characterized by real-valued non-negative weights $\{f_n\geq 0|n=0,1,2,\dots\}$ as the number of maximum possible number of hops away from non-zero prior values in which non-zero posteriors can be found before post-processing. We mathematically express this quantity as:
    $$\text{breadth}(diag(p)\mathcal{F}(\hat{A}))=\max_{n=0,1,2,\dots,dm}\{n:f_n\neq 0\}$$
    where $dm$ is the graph's unweighted diameter.
    As long as the adjacency matrix normalization does not assign  a zero value to any edge, the breadth does not depend on the exact type of normalization. Thus, for non-negative priors and any post-processing vector $p$, it holds that the conditions $\{r_{ns0}[v]=0,r_{ns}[v],r_0[0]=0,f_\delta(r)[v]=0\}$ can only simultaneously hold true for node $v$ (this does not necessarily hold true for $r[v]$, given that prior editing is allowed to procure negative values).

    For the rest of nodes, and given that $\delta+r_0[v]>0$, there exist appropriate values $r[v]=\tfrac{r_{ns}[v]}{r_{ns0}[v]}(\delta+r_0[v])-\delta$ corresponding to the optimal ones $r_{ns}[v]$. Importantly, for $r=r_0$ it also holds that $r_{ns0}=r_{ns}$, which means that the search around the original symmetric priors is translated to a search around the original asymmetric priors and not at a random area in $\mathbb{R}^{|\mathcal{V}|}$.

    As a final step, we rewrite the desired loss to be minimized as:
    $$\mathcal{L}_{ns}(r_{ns})=\mathcal{L}_{ns}\big(r_{ns0}\tfrac{\delta+r}{\delta+r_0}\big)=\mathcal{L}_{ns}\big(\tfrac{1}{\delta+r_0}r+\tfrac{r_{ns0}\delta}{\delta+r_0}\big)$$
    This is equivalent to applying the post-processing vector $p'=\tfrac{1}{\delta+r_0}p$ (where $p$ is the original post-processing) and adding a constant additive term to $r$ within the loss function. Given that the last additive term does not affect the supremum of loss gradients, our analysis is applicable as-is, with the only difference in our theoretical results being that attraction radii are equivalent to those that would have been calculated for post-processing vectors $\tfrac{p}{\delta+r_0}$.
\end{proof}

\end{document}